\documentclass[letterpaper,twocolumn,10pt]{article}
\usepackage{usenix}

\usepackage{algorithm}
\usepackage{atbegshi}
\usepackage[noend]{algpseudocode}
\usepackage{amsmath}
\usepackage{amssymb}
\usepackage{array}
\usepackage{cryptocode}
\usepackage{bm}
\usepackage{booktabs}
\usepackage{cite}
\usepackage{color}
\usepackage{dsfont}
\usepackage{enumitem}
\usepackage{eqparbox}
\usepackage{graphicx}
  \setkeys{Gin}{width=.49\textwidth, totalheight=\textheight, keepaspectratio}
  \graphicspath{{graphics/}}
\usepackage{hyperref}

\usepackage{nicefrac}
\usepackage{subcaption}
\usepackage{tikz}
	\usetikzlibrary{shapes}
	\usetikzlibrary{plotmarks}
\usepackage{xcolor}
\usepackage{xpatch}

\newcommand{\parabf}[1]{\vspace{1mm}\noindent\textbf{#1}}
\newcommand{\parait}[1]{\vspace{1mm}\noindent\textit{#1}}
\newcommand{\para}[1]{\vspace{1mm}\noindent#1}

\newcommand{\adv}{\mathcal{A}}
\newcommand{\advMIA}{\adv^{\mathcal{L}}}
\newcommand{\advAI}{\adv^\mathcal{I}}
\newcommand{\advantage}{Adv}
\newcommand{\advantageA}{\advantage^\adv}
\newcommand{\advantageAI}{\advantage^{\mathcal{I}}}
\newcommand{\advantageMIA}{\advantage^{\mathcal{L}}}
\newcommand{\advantageU}{\advantage^{\mathcal{U}}}
\newcommand{\advGame}{\mathcal{A}}
\newcommand{\chGame}{\mathcal{C}}
\newcommand{\coeffs}[1]{\mathbf{w}_{#1}}
\newcommand{\distPop}{\mathcal{D}_\mathcal{R}}
\newcommand{\distR}{\mathcal{D}_R}
\newcommand{\distGM}[1]{\mathcal{D}_{\GMtrained{#1}}}
\newcommand{\feature}{f}
\newcommand{\featureset}{\bm{f}}
\newcommand{\GMtrained}[1]{g(#1)}
\newcommand{\labeltest}{y_e}
\newcommand{\labelhat}{\hat{y}_e}
\newcommand{\nAttr}{k}
\newcommand{\nFeatures}{l}

\newcommand{\PG}{PG}
\newcommand{\pop}{\mathcal{R}}
\newcommand{\prior}{\mathcal{P}}
\newcommand{\Prob}[1]{\text{P}\left[#1\right]}
\newcommand{\PMtrained}[1]{h(#1)}
\newcommand{\PMtrainedX}[2]{h_{#1}(#2)}
\newcommand{\public}{b}
\newcommand{\record}{\bm{r}}
\newcommand{\secret}{s_t}
\newcommand{\secrethat}{\hat{s}_t}
\newcommand{\sensitive}{r_s}
\newcommand{\sensitivehat}{\hat{r}_s}

\newcommand{\sizeRadv}{l}
\newcommand{\sizeR}{n}
\newcommand{\sizeS}{m}
\newcommand{\synrecord}{\bm{s}}
\newcommand{\San}{\mathtt{San}}
\newcommand{\target}{\record_t}
\newcommand{\targetpartial}{\tilde{\target}}

\newcommand{\test}{\record_e}
\newcommand{\testpartial}{\tilde{\record}_e}
\newcommand{\trainGM}[1]{\mathtt{GM}(#1)} 
\newcommand{\trainPM}[1]{\mathtt{PM}(#1)}

\newcommand{\PrivacyAttack}{\mathtt{PrivacyAttack}}
\newcommand{\GenerativeModel}{\mathtt{GenerativeModel}}

\newcommand{\IndHist}{\mathtt{IndHist}}
\newcommand{\BayNet}{\mathtt{BayNet}}
\newcommand{\CTGAN}{\mathtt{CTGAN}}
\newcommand{\PrivBay}{\mathtt{PrivBay}}
\newcommand{\PATEGAN}{\mathtt{PATEGAN}}

\newcommand{\Fnaive}{\mathtt{F_{Naive}}}
\newcommand{\Fhist}{\mathtt{F_{Hist}}}
\newcommand{\Fcorr}{\mathtt{F_{Corr}}}

\newcommand{\Adult}{\texttt{Adult}}
\newcommand{\Texas}{\texttt{Texas}}

\tikzset{cross/.style={cross out, draw=black, minimum size=2*(#1-\pgflinewidth), inner sep=0pt, outer sep=0pt}, cross/.default={2pt}}
\newcommand\marksymbol[2]{\tikz[#2, scale=1.5pt]\pgfuseplotmark{#1};}

\newcommand{\tikzcircle}[1]{\tikz[baseline=-0.5ex]\draw[#1, fill=#1] (0,0) circle [radius=2pt];}%
\newcommand{\tikzcrossrot}[1]{\tikz[baseline=-0.5ex]\draw[line width=2pt] (0,0) node[cross=4pt, rotate=90, #1] {};}%
\newcommand{\tikzcross}[1]{\tikz[baseline=-0.5ex]\draw[line width=2pt] (0,0) node[cross=4pt, rotate=45, #1] {};}
\newcommand{\tikztriangle}[1]{\marksymbol{triangle*}{#1}}
\newcommand{\tikzdiamond}[1]{\marksymbol{diamond*}{#1}}
\newcommand{\tikzbar}[1]{\tikz[baseline=0ex]\draw[#1, fill=#1] (0,0) rectangle (0.5em,0.7em);}

\definecolor{safeBlue}{RGB}{136,204,238}
\definecolor{safeRed}{RGB}{204,102,119}
\definecolor{safeYellow}{RGB}{221, 204, 119}
\definecolor{safePurple}{RGB}{51,34,136}
\definecolor{safeGreen}{RGB}{17,119,51}
\definecolor{safePink}{RGB}{170,68,153}

\begin{document}

\title{Synthetic Data -- Anonymisation Groundhog Day}

\author{
{\rm Theresa Stadler}\\
EPFL
\and
{\rm Bristena Oprisanu}\\
UCL
\and
{\rm Carmela Troncoso}\\
EPFL
}

\maketitle

\begin{abstract}
\footnote{When citing this work, please note that a peer-reviewed version of this paper will be published at USENIX Security 2022. There are some minor editorial differences between the two versions.}
Synthetic data has been advertised as a silver-bullet solution to privacy-preserving data publishing that addresses the shortcomings of traditional anonymisation techniques. The promise is that synthetic data drawn from generative models preserves the statistical properties of the original dataset but, at the same time, provides perfect protection against privacy attacks.
In this work, we present the first quantitative evaluation of the privacy gain of synthetic data publishing and compare it to that of previous anonymisation techniques.

Our evaluation of a wide range of state-of-the-art generative models demonstrates that synthetic data either does \emph{not prevent inference attacks} or does not retain data utility. In other words, we empirically show that synthetic data does not provide a better tradeoff between privacy and utility than traditional anonymisation techniques.
Furthermore, in contrast to traditional anonymisation, the privacy-utility tradeoff of synthetic data publishing is hard to predict. Because it is impossible to predict what signals a synthetic dataset will preserve and what information will be lost, synthetic data leads to a highly variable privacy gain and unpredictable utility loss. In summary, we find that synthetic data is far from the holy grail of privacy-preserving data publishing.
\end{abstract}

\section{Introduction}
\label{sec:intro}

The rise of data-driven decision making as the prevailing approach to advance science, industrial production, and governance generates a need to share and publish data~\cite{EDS20, EWPAI20, FDS20}. At the same time, growing concerns about the implications that data sharing has for individuals and communities call for data publishing approaches that preserve fundamental rights to privacy.
%
Yet, how to share high-dimensional data in a privacy-preserving manner remains an unsolved problem. Attempts to anonymise micro-level datasets have failed across the board~\cite{Sweeney02, Ohm09, NarayananS09, NarayananSR11, deMontjoyeHVB13, deMontjoyeRSP15,  SweeneyY15, CulnaneRT18}. A large number of publications, case studies, and real-world examples demonstrate that high-dimensional, sparse datasets are inherently vulnerable to privacy attacks. The repeated failures to protect the privacy of microdata releases reflect a fundamental tradeoff: information-rich datasets that are valuable for statistical analysis also always contain enough information to conduct privacy attacks~\cite{NarayananS19}.

In this landscape, practitioners and researchers see in synthetic data a promising approach to open data sharing that addresses the privacy issues of previous anonymisation attempts~\cite{DrechslerR10, ChoiBMDSS17, XuSCV19 , YaleDDGPB19, YaleDDGPB19-A, BellovinDR19, ArnoldN20, SynAE19, ODI19, NIST19, SynAE20, SynUK20, Statice, Hazy}. Synthetic data is presented as ``the next, best step in sanitized data release''~\cite{BellovinDR19} that addresses a wide variety of privacy-sensitive use cases from deriving aggregate insights~\cite{RosenblattLPL20, SDGym} to outlier analysis~\cite{TuckerWRM20, MostlyAI_Fraud}. Synthetic datasets are promised to preserve the statistical properties of the original data but ``contain no personal data''~\cite{SynAE19} and hence ``enable the protection of personally identifiable information''~\cite{NIST19}. 
In this work, we present a rigorous, quantitative assessment of such claims and challenge the common perception of synthetic data as the holy grail of privacy-preserving data publishing. 

\parabf{Previous works.} Previous studies on the privacy properties of synthetic data publishing overestimate its benefits over traditional anonymisation for multiple reasons.
A common argument to support claims about the privacy benefits of synthetic data is that it is `artificial data' and therefore no direct link between real and synthetic records exists. Hence, many argue, synthetic data by design protects against traditional attacks on microdata releases such as linkage~\cite{Sweeney02, ElliotOROMDGPM18} or attribute disclosure~\cite{MachanavajjhalaGKV06, ElliotOROMDGPM18}.
Consequently, many studies rely on similarity tests between real and synthetic records to measure the privacy leakage of synthetic datasets~\cite{ChoiBMDSS17, YaleDDGPB19, YaleDDGPB19-A}. 
As we show in this paper, these studies severely underestimate the privacy risks of synthetic data publishing. We introduce two new privacy attacks that demonstrate that, despite its artificial nature, synthetic data does not protect all records in the original data from linkage and attribute inference.

More recent works analyse the vulnerability of generative models against model-specific extraction attacks~\cite{ChenYZF19, HayesMDC19, HilprechtHB19}. Due to their focus on white-box attacks against non-parametric models for synthetic image generation these works do not provide the right framework to assess the privacy risks of synthetic data sharing in the tabular data domain.
In contrast, our attacks treat the data synthesis method as a black-box, focus on tabular data publishing, and allow us to directly compare the privacy leakage of synthetic data to that of traditional anonymisation techniques. 

Other approaches rely on formal privacy guarantees for the generative model training process to prevent privacy attacks~\cite{AbowdV08, BindschaedlerSG17}.
While formal definitions of privacy are a clear improvement over the heuristic privacy models of traditional anonymisation, their guarantees are often hard to interpret and difficult to compare to alternative anonymisation techniques.
Here, we propose a framework that enables data holders to empirically evaluate the privacy guarantees of differentially private synthetic data publishing and to directly compare its tradeoffs to that of traditional anonymisation techniques.

\parabf{Contributions.} In this paper, we \emph{quantitatively assess} whether (differentially private) synthetic data produced by a wide range of common generative model types does provide a higher gain in privacy than traditional sanitisation at a lower cost in utility. Our results demonstrate that: 

\noindent (I) Synthetic data drawn from generative models without explicit privacy protection \emph{does not protect outlier} records from linkage attacks. Given access to a synthetic dataset, a strategic adversary can infer, with high confidence, the presence of a target record in the original data.

\noindent (II) Differentially private synthetic data that hides the signal of individual records in the raw data protects these targets from inference attacks but does so \emph{at a significant cost in utility}. Worse, in contrast to traditional anonymisation techniques, synthetic datasets do not give any transparency about this tradeoff. It is impossible to predict what data characteristics will be preserved and what patterns will be suppressed.

\noindent (III) Our empirical evaluation of the existing implementations of two popular differentially private generative model training algorithms reveals that certain implementation decisions \emph{violate their formal privacy guarantees} and leave some records vulnerable to inference attacks. We provide a novel implementation of both algorithms that addresses these shortcomings.

\noindent (IV) We make our evaluation framework available as an \emph{open-source library}. Our implementation allows practitioners and researchers to quantify the privacy gain of publishing a synthetic in place of a raw or sanitised dataset and compare the quality of different anonymisation mechanisms. The framework includes implementations of two relevant privacy attacks and can be applied to any type of generative model training algorithm.


\section{Synthetic data and generative models}
\label{sec:syntheticdef}

In this section, we formalise the process of synthetic data generation. \autoref{tab:notation} in the Appendix summarises our notation.
Let $\pop$ be a population of data records where each record $\record \in \pop$ contains $\nAttr$ attributes: $\record = (r_1, \cdots, r_\nAttr)$. We denote the unknown joint probability distribution over the data domain of the population as $\distPop$. We refer to $R \sim \distPop^\sizeR$, a collection of $\sizeR$ data records sampled independently from $\distPop$, as raw dataset which defines the data distribution $\distR$.

\parabf{Synthetic data generation.} The goal of a generative model is to learn a representation of the joint probability distribution of data records $\distR$. 
The model training algorithm $\trainGM{R}$ takes as input a raw dataset $R$, learns $\distGM{R}$, a representation of the joint multivariate distribution $\distR$, and outputs a trained generative model $\GMtrained{R}$. The model $\GMtrained{R}$ is a stochastic function that, without any input, generates synthetic records $\synrecord_i$, distributed according to $\distGM{R}$. We denote the process of sampling a synthetic dataset $S = (\synrecord_1, \cdots, \synrecord_\sizeS)$ of size $\sizeS$ as $S \sim \distGM{R}^\sizeS$.
We write $\GMtrained{R} \sim \trainGM{R}$ instead of $\GMtrained{R} \gets \trainGM{R}$ to indicate that the training algorithm can be a stochastic and non-deterministic process.
 
\parabf{Approximation by features.} It is tempting to assume that the model $\distGM{R}$ provides a perfect representation of the data distribution $\distR$ and that synthetic data ``carries through all of the statistical properties, patterns and correlations in the [input] data''~\cite{HazyIntro}. The trained model, however, only provides a \emph{lower-dimensional approximation} of the true data distribution. It retains \emph{some} characteristics but can never preserve all of them. Which characteristics are captured, and how they approximate $\distR$, is determined by the generative model choice.
\emph{Statistical models}, such as Bayesian networks \cite{KollerF09}, or Hidden Markov models~\cite{Ghahramani01}, provide an explicit, parametric model of $\distR$. The features these models extract from their training data is determined upfront.
\emph{Non-parametric models}, such as generative adversarial networks (GANs)~\cite{GoodfellowPMXWOCB14} or variational auto encoders (VAEs)~\cite{KingmaMRW14}, do not estimate a parametric likelihood function to generate new samples from $\distGM{R}$. Which features of the input data are most relevant and how the model approximates $\distR$ is implicitly determined during training~\cite{Goodfellow16}. 

The features a generative model uses to approximate $\distR$ define which of the statistical properties of the raw data $R$ are replicated by a synthetic dataset $S \sim \distGM{R}$ sampled from the trained model.
Statistical models provide some control over what features will be preserved. However, it is not possible to \emph{exclude} that a synthetic dataset reproduces characteristics of the original data other than the features explicitly captured by the model. For instance, synthetic data generated through independent sampling from a set of 1-way marginals is expected to preserve a dataset's independent frequency counts. However, if the raw data contains strong correlations between attributes, these correlations are likely to be replicated in the synthetic data even under independent sampling.

\subsection{Generative models in this study}\label{subsec:GMs}
In Sections \ref{subsec:mia_results} and \ref{subsec:put_attribute}, we empirically evaluate the privacy gain of synthetic data publishing for five existing generative model training algorithms.
We implemented three generative models without explicit privacy protection and two models with differential privacy guarantees. We chose models relevant to the tabular data sharing use case and to cover a wide range of model architectures. We further considered their computational feasibility for high-dimensional datasets and whether a working implementation was available. \autoref{tab:modelparams} in Appendix \ref{ap:implementation} lists our parametrisation of these models.

\para{$\IndHist$.} The $\IndHist$ training algorithm from Ping et al.~\cite{PingJB17} extracts marginal frequency counts from each data attribute and generates a synthetic dataset $S$ through independent sampling from the learned marginals. Continuous attributes are binned. The number of bins is a configurable model parameter.

\para{$\BayNet$.} Bayesian networks capture correlations between attributes by factorising the joint data distribution as a product of conditionals. The degree of the network model is a model parameter. The trained network provides an efficient way to sample synthetic records from the learned distribution (see Zhang et al.~\cite{ZhangCPSX17} for details). We use the GreedyBayes implementation provided by Ping et al.'s DataSynthesizer~\cite{PingJB17}.

\para{$\PrivBay$.} PrivBayes~\cite{ZhangCPSX17} is a differentially private Bayesian network model. Both, the Bayesian network and the conditional distributions, are learned under $\varepsilon$-differentially private algorithms. A synthetic dataset can be sampled from the trained model without any additional privacy budget cost. We use the GreedyBayes procedure provided by Ping et al.~\cite{PingJB17} to train a differentially private version of $\BayNet$ such that a $\PrivBay$-trained model with $\varepsilon \rightarrow \infty$ corresponds to a $\BayNet$-model without formal guarantees.

\para{$\CTGAN$.} CTGAN~\cite{XuSCV19} uses mode-specific normalisation of tabular data attributes to improve the approximation of complex distributions through GANs. CTGAN further uses a conditional generator and training-by-sampling to get better performance on imbalanced datasets.

\para{$\PATEGAN$.} PATEGAN builds on the Private Aggregation of Teacher Ensembles (PATE) framework~\cite{PapernotAEGT17} to achieve DP for GANs~\cite{JordonYS19}. PATEGAN replaces the discriminator's training procedure with the PATE mechanism. The trained model provides $(\varepsilon, \delta)$-DP with respect to the discriminator's output.

\section{Quantifying the privacy gain of synthetic data publishing}\label{sec:framework}
The promise of synthetic data is that it allows data holders to publish (synthetic) datasets that are useful for analysis while, at the same time, protect the privacy of individuals in the raw data against powerful adversaries~\cite{DrechslerR10, ChoiBMDSS17, XuSCV19 , YaleDDGPB19, YaleDDGPB19-A, BellovinDR19, ArnoldN20}. The increasing number of applications of synthetic data tools shows that this has become an appealing proposition~\cite{SynAE19, ODI19, NIST19, SynAE20, SynUK20, Statice, Hazy}.
Here, we introduce a novel evaluation framework that allows data holders to quantitatively assess claims about the privacy benefits of synthetic data sharing.

\parabf{Synthetic data as an anonymisation mechanism.} Synthetic data providers often present synthetic data as a novel ``data anonymisation solution''~\cite{Statice} that addresses the shortcomings of traditional sanitisation techniques, such as generalisation~\cite{Sweeney02, MachanavajjhalaGKV06} or perturbation~\cite{NarayananS08}. Data holders are promised that publishing a synthetic in place of the raw dataset prevents the leakage of private information about individuals in the raw data previously observed in sanitised datasets~\cite{Sweeney02,NarayananS08}. 

To evaluate this claim, that synthetic data generation is an effective anonymisation mechanism, we hence need to assess whether synthetic data addresses the privacy risks that originally motivated the use of data anonymisation techniques. These are the risk of \emph{linkability} and \emph{inference}~\cite{Art29WP}.
Previous anonymisation methods 
such as k-anonymity~\cite{Sweeney02} or l-diversity~\cite{MachanavajjhalaGKV06} have failed to provide robust protection against these attacks for high-dimensional, sparse datasets~\cite{NarayananS08}. So far, however, there is no evidence that synthetic data provides \emph{better protection} against these attacks at a lower cost in utility than traditional sanitisation techniques.

A number of recent papers have tackled related problems but focus primarily on non-parametric models for synthetic image generation and adversaries with white-box or query access to the model~\cite{ChenYZF19, HayesMDC19, HilprechtHB19}. However, like sanitisation, synthetic data is primarily seen as a tool for privacy-preserving tabular data sharing, i.e., to enable data holders to publish a single copy of synthetic data as opposed to the trained model or a set of statistics~\cite{SynAE20, NIST19, SynUK20, Hazy, Statice}. In our framework, we hence assume that the adversary only has access to a synthetic dataset and can not repeatedly query the trained model or observe its parameters.

\subsection{Evaluation framework}
The goal of our framework is to quantitatively assess whether publishing a synthetic dataset $S$ in-place of the raw data $R$ reduces the privacy risks for individuals in the raw data with respect to the relevant privacy concerns.
We model each privacy concern as an adversary $\adv$ that given a raw or synthetic dataset aims to infer a secret about a target record $\target$ from the population $\pop$. For each adversary, we define an advantage measure $\advantageA$ that captures by how much including an individual's record in the published data increases this individual's privacy risk. In \autoref{subsec:mia_game} and \ref{subsec:put_attribute}, we define adversaries and advantage measures that model the risk of linkability and inference, respectively.

\parabf{Privacy gain.} We assess the privacy gain of publishing a synthetic dataset $S$ in place of the raw data $R$ for target record $\target$ as the reduction in the adversary's advantage when given access to $S$ instead of $R$

\begin{equation}\label{eq:gain}
	\PG \triangleq \advantageA \left( R, \target \right) - \advantageA \left( S, \target \right).
\end{equation} 

A high privacy gain indicates that publishing $S$ in place of $R$ substantially reduces the privacy risk modelled by adversary $\adv$ for target record $\target$. 
A low gain, in contrast, implies that the data holder's decision to publish $S$ or $R$ has no impact on the privacy loss for target record $\target$, i.e., the adversary's advantage remains the same.

The privacy gain hence allows us to assess whether synthetic data is, as promised, an effective anonymisation mechanism. A good anonymisation mechanism should result in a high privacy gain for all records in the population and under any potential privacy adversary. A low gain in privacy indicates that the anonymisation mechanism does not provide \emph{a significant improvement over publishing the raw data}.

\parabf{Comparison to previous evaluation approaches.}
In contrast to model-specific evaluation techniques~\cite{HuRW14}, our framework treats the data generating mechanism as a \emph{complete black-box} and evaluates the privacy risks of \emph{synthetic data publishing} rather than an adversary's inference power when given query or white-box access to a model~\cite{ReiterWZ14, HayesMDC19}.
As opposed to privacy evaluations based on similarity metrics~\cite{ChoiBMDSS17, YaleDDGPB19, YaleDDGPB19-A}, the framework provides data holders with a direct measure of how well the synthetic data defends against the privacy risks of data sharing. This makes the evaluation results easily interpretable and relatable to relevant data protection regulations~\cite{Art29WP}.
We design and implement the framework in a modular fashion. This ensures that the framework is not limited to a specific threat model~\cite{ReiterWZ14} or the privacy risks modelled in this paper. Instead, it can be adapted to any privacy concern specific to the data holder's use case. The proposed evaluation method is independent of the data generation method. Thus, it can be used to evaluate the privacy gain of synthetic data generated by models trained without any explicit privacy protection, models trained under formal privacy guarantees~\cite{AbowdV08, BindschaedlerSG17}, or traditional anonymisation techniques.

\parait{Worst-case vs. average-case evaluation.} Finally, previous studies have shown that the privacy risks of data sharing are not uniformly distributed across the population~\cite{YaghiniKCT19, RocherHM19, LongWB20}. While individuals that are representative of a large majority of the population are often protected from privacy attacks, outliers or members of minorities largely remain vulnerable. Our framework allows to assess privacy risks both at an aggregate population-level and on a per-record basis. This enables us to demonstrate that synthetic data provides disparate privacy gain across population subgroups.

\section{Does synthetic data mitigate the risk of linkability?}\label{sec:mia}
A major privacy concern in the context of privacy-preserving data sharing is the risk of linkability. Linkage attacks aim to link a target record to a single record, or group of records, in a sensitive dataset. Linkage enables adversaries to attach an identity to a supposedly de-identified record~\cite{Sweeney02, NarayananS08} or to simply establish the fact that this particular record is present in a sensitive dataset~\cite{HomerSRDTMPSNC08}.

\parabf{Related work.} The risk of linkability has been demonstrated, in theory and practice, for a large variety of data types: tabular micro-level datasets~\cite{Sweeney02, NarayananS08}, social graph data~\cite{NarayananS09, NarayananSR11}, aggregate statistics~\cite{PyrgelisTD18}, and statistical models~\cite{ShokriSSS17}.
Linkage attacks on tabular microdata usually intend to link a target record (connected to an identity) to a single record in a sensitive database from which direct identifiers have been removed.

Membership inference attacks (MIAs) are linkage attacks which target the output of statistical computations run on sensitive datasets, such as aggregate statistics~\cite{HomerSRDTMPSNC08, PyrgelisTD18} or trained ML models~\cite{ShokriSSS17}. ML-oriented MIAs have been extensively studied on predictive models, such as binary or multi-label classifiers~\cite{ShokriSSS17, YeomGFJ18, SalemZHBFB18, LeinoF19}. Recently, this work has been extended to GANs and VAEs~\cite{HayesMDC19, ChenYZF19, HilprechtHB19}.

\subsection{Formalizing linkability as membership inference}\label{subsec:mia_game}
In a linkage attack, the adversary aims to learn whether a record is present in a sensitive dataset. Following works by Yeom et al.~\cite{YeomGFJ18} and Pyrgelis et al.~\cite{PyrgelisTD18}, we hence model the risk of linkability as a membership privacy game between an adversary $\advGame$ and a challenger $\chGame$. The challenger plays the role of a data holder that publishes a dataset $X$ that is made available to the adversary. This dataset could either be a raw dataset $R$ or a sanitised or synthetic version of $R$, denoted as $S$. The goal of the adversary $\advGame$ is to infer whether a target record $\target$, chosen by the adversary, is present in the sensitive dataset $R$ based on the published dataset $X$ and some prior knowledge $\prior$.

\begin{figure}[h]
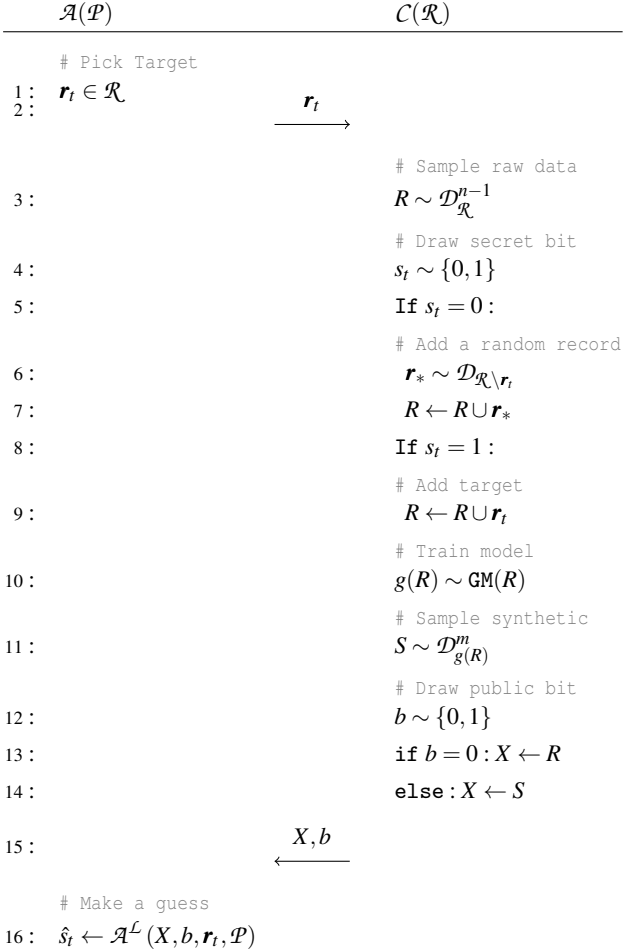

\newcommand{\comment}[1]{\textcolor{gray}{\footnotesize{\texttt{\# #1}}}}
 \pseudocodeblock[colsep=0em,skipfirstln, linenumbering]{
 	\pcskipln
 	\advGame(\prior) \< \< \chGame(\pop) \\[0.1\baselineskip][\hline]\pcskipln
 	\< \< \\[-0.5\baselineskip]
	\comment{Pick Target}\< \<\\[-0.2\baselineskip]
 	\target \in \pop\< \< \\[-1\baselineskip] \pcskipln
 	\< \sendmessageright*[1cm]{\target} \<\\
	\< \< \comment{Sample raw data}\\[-0.2\baselineskip]  \pcskipln
 	\< \< R \sim \distPop^{\sizeR-1}\\
	\< \< \comment{Draw secret bit}\\[-0.2\baselineskip]
 	\< \< \secret \sim \{0,1\}\\ \pcskipln
 	\< \< \mathtt{If}~\secret=0:\\ 
	\< \< \comment{Add a random record}\\[-0.2\baselineskip]
 	\< \< \;\;\record_* \sim \mathcal{D}_{\pop \setminus \target}\\
 	\< \< \;\;R \gets R \cup \record_*\\ \pcskipln
 	\< \< \mathtt{If}~\secret=1:\\
	\< \< \comment{Add target}\\[-0.2\baselineskip] \pcskipln
 	\< \< \;\;R \gets R \cup \target\\
	\< \< \comment{Train model}\\[-0.2\baselineskip] \pcskipln
 	\< \< \GMtrained{R} \sim \trainGM{R}\\
	\< \< \comment{Sample synthetic}\\[-0.2\baselineskip] \pcskipln
 	\< \< S \sim \distGM{R}^\sizeS\\
	\< \< \comment{Draw public bit}\\[-0.2\baselineskip]
 	\< \< \public \sim\{0,1\}\\
	\< \< \mathtt{if}~\public = 0: X \gets R\\
	\< \< \mathtt{else}: X \gets S\\ \pcskipln
 	\< \sendmessageleft*[1cm]{X, \public} \<\\
	\comment{Make a guess}\< \<\\[-0.2\baselineskip]
 	\secrethat \gets \advMIA\left( X, \public, \target, \prior \right)\<\<%
 	}
\caption{Linkability privacy game.}\label{fig:mia_game}
\end{figure}

\autoref{fig:mia_game} presents the linkability game for the case where $S$ is a synthetic dataset sampled from a generative model trained on $R$. Later, we discuss how the challenger's protocol changes when the game models sanitisation.
First, $\advGame$ picks a target record $\target$ and sends it to $\chGame$. $\chGame$ draws a raw dataset $R$ of size $\sizeR - 1$ from the distribution defined by the population $\pop$, and a secret bit $\secret \sim \{0, 1\}$. 
If $\secret = 0$, $\chGame$ draws a random record $\record_*$ from the population (excluding the target) and adds it to the raw dataset.
If $\secret = 1$, $\chGame$ adds the target $\target$ to the raw dataset. 
Then, $\chGame$ trains a generative model on the raw data $R$, and samples a synthetic dataset $S$ of size $\sizeS$ from the trained model.
$\chGame$ picks at random whether to send back to the adversary the raw data $R$ or the synthetic data $S$. 
$\advGame$ receives the dataset and makes a guess about the target's presence in $R$, $\secrethat \gets \advMIA \left(X, \public, \target, \prior \right)$. The adversary wins the game if $\secrethat = \secret$.     

As in Yeom et al.~\cite{YeomGFJ18}, we assume an equal prior over the target's membership in $R$ and define the linkage adversary's advantage as:

\begin{align}\label{eq:mia_adv}
	\advantageMIA \left(X, \target \right) &\triangleq 2\Prob{\advMIA \left(X, \public, \target, \prior \right) = \secret} - 1\\
	&= \Prob{\secrethat = 1 | \secret = 1} - \Prob{\secrethat = 1 | \secret = 0}
\end{align}

where $X$ can be a raw $R$ or synthetic dataset $S$. The probability space of $\advantageMIA$ is defined by the random choices of $R \sim \distPop$ and $\secret \sim \{0, 1\}$ and the randomness of the synthetic data generation mechanism and the adversary's guess.

\parabf{Adversarial strategy.} The adversary's guess function $\advMIA(\cdot)$ takes as input a target record $\target$, the information published by the challenger, $X$ and $\public$, and some prior information $\prior$ and outputs a guess about the target's presence in $R$. The adversary's strategy to make a guess changes depending on the data published.\\
If $\chGame$ publishes $X = R$, the adversary simply checks whether $\target \in R$ and has a probability of $1$ to win the game ($\advantageMIA(R, \target) = 1$).\\
If $X = S$, the adversary performs a binary classification task on a set of features extracted from the synthetic data $S$. We explain how we implement this binary classifier in \autoref{subsec:mia_implementation}.

\parabf{Privacy gain.} For each target chosen by the adversary, we instantiate the game multiple times and measure the adversary's advantage conditioned on the challenger's choice of $X$. Under our definition of privacy gain in \autoref{eq:gain} and with $\advantageMIA(R, \target) = 1$, the privacy gain of publishing a synthetic dataset $S$ in place of the raw data with respect to the risk of linkability is given as:

\begin{equation}
	\PG = 1 - \advantageMIA(S, \target)
\end{equation}                       

A privacy gain of $\PG = 0$ indicates that the adversary infers the target's presence in $R$ with perfect accuracy regardless of whether given access to the raw or synthetic data. If on the other hand, observing the synthetic data $S$ gives the adversary no advantage in inferring the target's presence ($\advantageMIA(S, \target) = 0$), then $\PG = 1$.

\subsection{A black-box membership inference attack}\label{subsec:mia_implementation}
We implement the adversary's strategy as a \emph{generic black-box} MIA that is independent of the generative model architecture. 

\parabf{Related work.} Existing MIAs on generative models focus almost exclusively on non-parametric deep learning models for synthetic image generation~\cite{HayesMDC19, ChenYZF19, HilprechtHB19, MukherjeeXTP21}. These works mostly investigate the privacy risks of either model-specific \emph{white-box} attacks or \emph{set membership} attacks that assume the adversary has access to the entire universe of training records and come to the conclusion that black-box MIAs that target specific records perform only slightly better than random baseline guessing~\cite{HayesMDC19, HilprechtHB19}. Unfortunately, previous attacks do not provide a good basis to evaluate the privacy gain of synthetic data publishing. Non-parametric models for non-tabular data cover only a very small set of use cases~\cite{SynAE20, NIST19, Hazy, Statice}, white-box attacks do not adequately reflect the data sharing scenario, and set inference attacks are not suitable to assess individual-level privacy gain.

\parabf{Shadow model attack.} In order to win the linkability game (see~\autoref{fig:mia_game}) when she receives a synthetic dataset $S$, the adversary needs a distinguishing function $\advMIA(\cdot)$ that enables her to infer the membership of $\target$ in the raw data $R$ used to train the generative model that output $S$. As in many previous works, we cast membership inference as a supervised learning problem and instantiate the adversary's guess function with a machine learning classifier trained on data produced by generative shadow models~\cite{ShokriSSS17, PyrgelisTD18}.

As Shokri et al.~\cite{ShokriSSS17}, we assume that, as part of her prior knowledge $\prior$, the adversary has access to the training algorithm $\trainGM{\cdot}$, the size of the raw and synthetic datasets $\sizeR$ and $\sizeS$, and to a reference dataset $R_\adv \sim \distPop^{\sizeRadv}$ that comes from the same distribution as the target model's training data $R \sim \distPop^\sizeR$ and may or may not overlap with $R$.
Given this prior knowledge $\prior$ and a target record $\target$, the adversary uses the following procedure to learn $\advMIA$: First, the adversary samples multiple training sets $R_i$ of size $\sizeR$ from the reference dataset $R_\adv$. On each set $R_i$, the adversary trains a generative model $\GMtrained{R_i}$ using the training procedure $\trainGM{R_i}$. From each of the trained models, the adversary samples multiple synthetic datasets $S$ of size $\sizeS$ and assigns them the label $\secret = 0$. The adversary repeats the same procedure on the same training sets, this time including the target, $R'_i = R_i \cup \target$, and assigns the generated synthetic datasets the label $\secret = 1$. Finally, the adversary trains a classifier $\advMIA$ on the labelled datasets. The trained classifier takes as input a synthetic dataset $S$ and outputs a guess $\secrethat$ about the target's presence in $R$: $\secrethat \gets \advMIA \left( S, \target, \prior \right)$.

\parabf{Feature extraction.} Existing MIAs on predictive models leverage patterns in the confidence values output by a trained model that differ between two classes, members and non-members~\cite{ShokriSSS17}. Mounting a successful black-box MIA on a generative model is much more challenging~\cite{HayesMDC19}. The attacker needs to identify the influence that \emph{a single target record} has on the high-dimensional data distribution $\distGM{R}$ as opposed to a low-dimensional confidence vector. Moreover, the output sampling process introduces additional uncertainty and the adversary only has access to a single output example.

In other words, the adversary needs to be able to distinguish between two distributions, $\distGM{R \cup \record_*}$ and $\distGM{R \cup \target}$, given a single observation $S \sim \distGM{X}^\sizeS$. To reduce the effect of high-dimensionality and sampling uncertainty, the adversary can apply \emph{feature extraction techniques}. Instead of training a classifier directly on $S$, the adversary learns to distinguish feature vectors extracted from synthetic datasets produced by models trained with and without the target, respectively.
A feature set can be described as a function $\feature(X) = \featureset$ that takes as input a set of records $X$ from the high-dimensional data domain and outputs a numerical vector $\featureset$ that maps $X$ into a lower-dimensional feature space.
Whether the attack using feature set $\featureset$ is successful depends on two factors: First, whether the target's presence has a detectable impact on any of the features in $\featureset$, and second, whether the synthetic dataset has preserved these features from the raw data and hence preserved the target's signal.

\parabf{Implementation.} We implement the distinguisher function $\advMIA$ as an instantiation of our framework's $\mathtt{PrivacyAttack}$ class (see Appendix~\ref{ap:implementation}). We leverage the object-oriented structure of the library to create multiple attack versions that share the same training procedure but use different attack models and feature extraction techniques.
As feature extractors, we implemented a naive feature set with simple summary statistics $\Fnaive$, a histogram feature set that contains the marginal frequency counts of each data attribute $\Fhist$, and a correlations feature set that encodes pairwise attribute correlations $\Fcorr$ (see Appendix~\ref{ap:implementation}).

As attack models, we implemented a Logistic Regression, Random Forests and K-Nearest Neighbours classifier. All attack models yielded similar results with a Random Forests classifier with 100 estimators using the Gini impurity splitting criterion performing best across datasets, generative models, and feature sets. In the remainder of the paper we focus on results obtained using this classifier.

\subsection{Empirical evaluation}\label{subsec:mia_results}
We first evaluated the expected privacy gain with respect to the risk of linkability under the three generative models trained without any formal privacy (see \autoref{subsec:GMs}) on two common benchmark datasets: \Adult\ and \Texas. Both are tabular datasets that contain a mix of numerical and categorical attributes. A detailed description of their characteristics can be found in Appendix \ref{ap:datasets}.

\parabf{Experiment procedure.} We aim to assess whether synthetic data produced by a wide range of generative model types does, as claimed, provide robust protection against linkage attacks. If synthetic data is a ``valid, privacy-conscious alternative to raw data''~\cite{BellovinDR19}, then its privacy gain should be close to $\PG = 1$ for all target records regardless of the attacker's strategy. We evaluate two groups of targets: five records randomly chosen from the population and five manually chosen outlier records representative of population minorities and most likely to be vulnerable to linkage attacks~\cite{NarayananS08}. As outliers, we selected records that either have rare categorical attribute values or numerical values outside the attribute's $95\%$ quantile. For instance, in the \Texas\ dataset we show the privacy gain for two records that have high total charges and one record with high total non-covered charges outside the attribute's $95\%$ quantile, and two records with an unusually high risk mortality and illness severity. For the \Adult\ dataset we followed the same procedure to select outlier records.

At the beginning of each experiment, we sample a fixed reference dataset $R_\adv$ of size $\sizeRadv$ from the population and use it to train the adversary's distinguisher. For each target record, we train multiple attack models using the shadow model training procedure described in \autoref{subsec:mia_implementation}.
To assess privacy gain, we repeatedly instantiate the linkability game described in \autoref{fig:mia_game} for each of our ten targets.

\begin{figure}
	\includegraphics[width=0.5\textwidth]{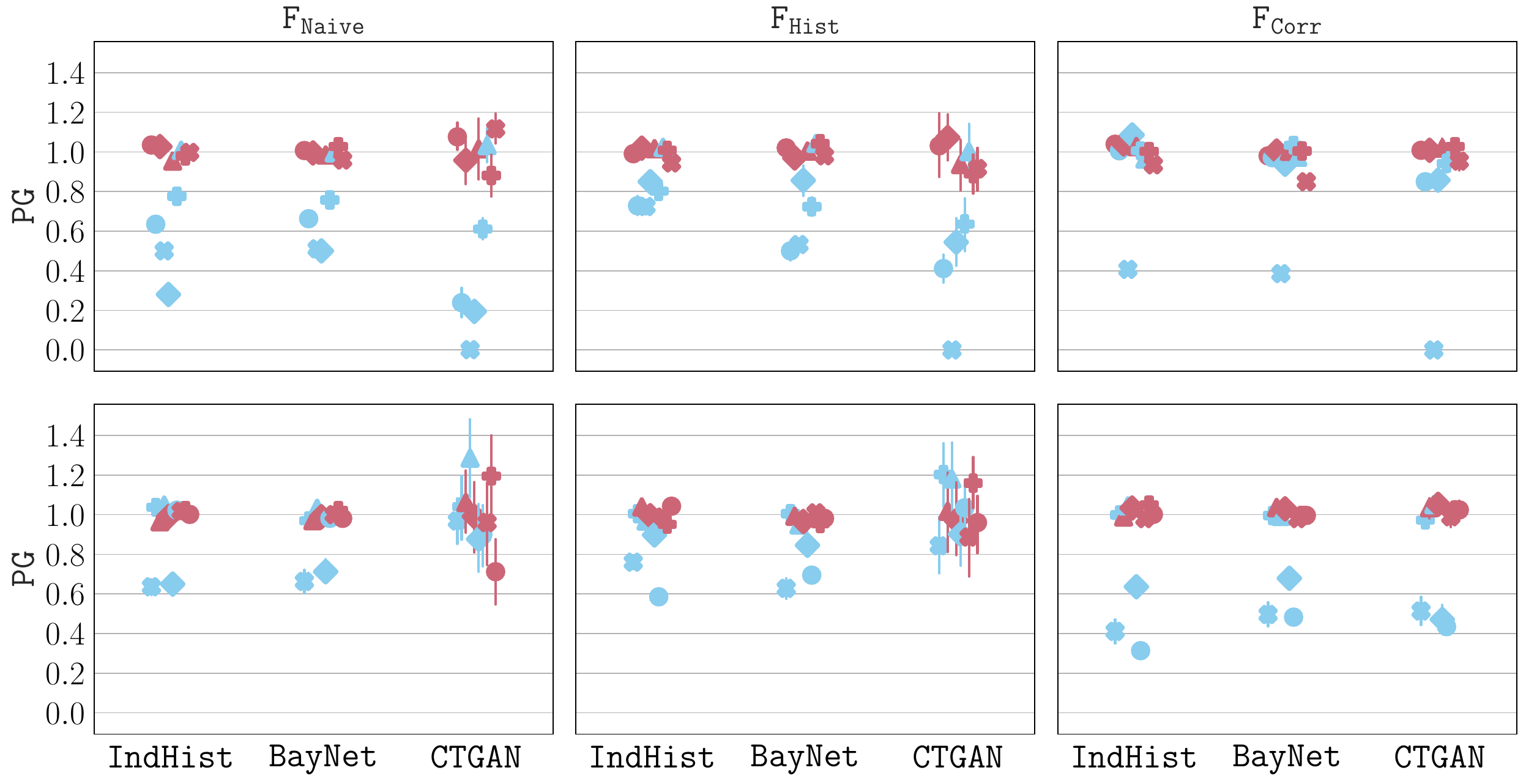}
	\caption{Expected per-record privacy gain for outliers~\protect\tikzcircle{safeBlue} and random records~\protect\tikzcircle{safeRed} for the \Texas~(\textit{top row}) and \Adult~(\textit{bottom row}) datasets under three different attacks using three distinct feature sets. Error bars represent the standard deviation.}
	\label{fig:mia_gain_targets}
\end{figure}

\parabf{Disparate gain.} \autoref{fig:mia_gain_targets} shows the average privacy gain across multiple instantiations of the linkability game for five outlier targets~\tikzcircle{safeBlue} and five randomly chosen targets~\tikzcircle{safeRed} for the \Texas~(top row) and \Adult~(bottom row) datasets, respectively. Each dataset, raw and synthetic, contained $\sizeR=\sizeS=1000$ records. The adversary was trained on a reference dataset of $\sizeRadv=10,000$ records using $10$ shadow models.  

We find that in both datasets privacy gain is unevenly distributed across target records. While the control group of randomly chosen target records (\tikzcircle{safeRed}) achieves close to perfect protection ($\PG \approx 1$), other records (\tikzcircle{safeBlue}) remain highly vulnerable to our linkage attack. The privacy gain for the majority of outlier targets is substantially smaller than $\PG=1$ (ideal case). For instance, under an attack using the naive feature set $\Fnaive$ $4$ out of the $5$ selected targets in the \Texas\ dataset achieve an average gain smaller than $0.8$ across all three generative models.
More worryingly, $1$ out of the $5$ targets tested (\tikzcrossrot{safeBlue}) consistently receives a privacy gain close to $0$ ($\PG < 0.005$) from synthetic data produced a by $\CTGAN$-trained model.
These results indicate that, contrary to claims by previous works~\cite{HayesMDC19}, publishing the synthetic in place of the raw data does not protect outlier targets from linkage attacks. 

\parabf{Unpredictable gain.} Which records remain at risk varies across generative model type and the adversary's feature set. In the \Texas\ dataset, an attack using the $\Fnaive$ feature set on $\CTGAN$-produced synthetic data results in a privacy gain below $\PG < 0.3$  for $3$ out of the $5$ outlier targets (\tikzcircle{safeBlue}, \tikzdiamond{safeBlue}, \tikzcrossrot{safeBlue}). The same attack on the same targets is less effective on synthetic data produced by $\IndHist$-trained models. The same group of targets reaches a maximum gain of $\PG = 0.77$.
Attacks on the \Adult\ dataset are most successful under the correlations feature set $\Fcorr$. Here, $\IndHist$-trained models provide a minimum gain of $\PG = 0.64$ (\tikzcrossrot{safeBlue}) under $\Fnaive$. The minimum gain provided by the same model drops below $\PG < 0.32$ if the attacker uses $\Fcorr$ as input to the attack and leaves a different target (\tikzcircle{safeBlue}) most vulnerable.

\parabf{Conclusions.} These results are extremely problematic from the point of view of a data holder seeking to use synthetic data generation as a privacy mechanism. Ideally, data holders should be able to predict, given a fixed dataset $R$ and a generative model training algorithm $\trainGM{\cdot}$, the minimum gain in privacy they can achieve. 
Our experiment shows, however, that this is next to impossible: The level of protection a generative model provides depends on how much information the model’s output leaks about the features targeted by the attack. This means that we can only predict privacy gain if we can (1) predict what features a potential adversary will target and (2) whether the synthetic data has preserved these features from the raw data.
In practice, neither of these factors is predictable. First, like traditional linkage attacks on microdata releases, a strategic adversary might use any set of features that are likely to be influenced by the target's presence~\cite{NarayananS08}. Second, which characteristics a synthetic dataset might preserve is not constrained to the features explicitly represented by the model. For instance, even the simplest statistical model $\IndHist$ might unexpectedly preserve features targeted by the attack: Synthetic data produced through independent attribute sampling by an $\IndHist$-model trained on the \Adult\ dataset leaves some target records vulnerable to linkage attacks using the correlations feature set $\Fcorr$. Non-parametric models, such as $\CTGAN$, do not even provide a parametric specification for the data’s density function. This makes it even harder to predict what set of features the model will preserve and an attack might target.

Previous assessments of the privacy risks of synthetic data publishing based on aggregate population measurements severely underestimate the risk of linkage attacks~\cite{HayesMDC19, HilprechtHB19 , GambsLLR21}. Our experimental evaluation reveals that synthetic data does not provide uniform protection against strategic adversaries, and some outliers remain highly vulnerable.    

\section{Does differentially private synthetic data mitigate the risk of linkability?}\label{sec:dp}

In the previous section, we demonstrate that non-private data synthesis algorithms are largely unsuitable as privacy mechanisms. This is not an unexpected finding as none of the evaluated models were originally designed as anonymisation mechanisms.  
In this section, we thus extend our analysis to two model training algorithms explicitly designed to protect the privacy of a model's training set, $\PrivBay$ and $\PATEGAN$. We evaluate to which extent their formal privacy guarantees improve the privacy gain of synthetic data publishing with respect to the risk of linkability.

\parabf{Differentially private generative models.} Model training algorithms based on the differential privacy model protect the privacy of the training data through formal guarantees for the \emph{lower-dimensional approximation of the full-dimensional data distribution}~\cite{ZhangCPSX17, BindschaedlerSG17, JordonYS19, GambsLLR21}. Synthetic datasets drawn from differentially private models preserve these privacy guarantees under the post-processing guarantee~\cite{DworkMNS06}.
 
The model training algorithm $\PrivBay$ learns a differentially private Bayesian network that approximates the relationship between data attributes via the exponential mechanism and computes the conditionally independent marginals in the subspaces of the Bayesian network via the Laplace Mechanism~\cite{ZhangCPSX17}.
$\PATEGAN$, a differentially private GAN, ensures that the discriminator's decisions are not affected by the presence of a single record in the model's training set by more than the defined $\varepsilon$-bound~\cite{JordonYS19}. 

\subsection{Empirical evaluation}\label{subsec:dp_results}
We used the experimental procedure described in \autoref{subsec:mia_results} to evaluate the privacy gain of $\PrivBay$ and $\PATEGAN$.
We integrated the implementations of these algorithms provided by Ping et al.~\cite{PrivBay} and Jordon et al.~\cite{PATEGAN} into our framework and ran the linkability game defined in \autoref{fig:mia_game}.

\begin{figure}
	\includegraphics{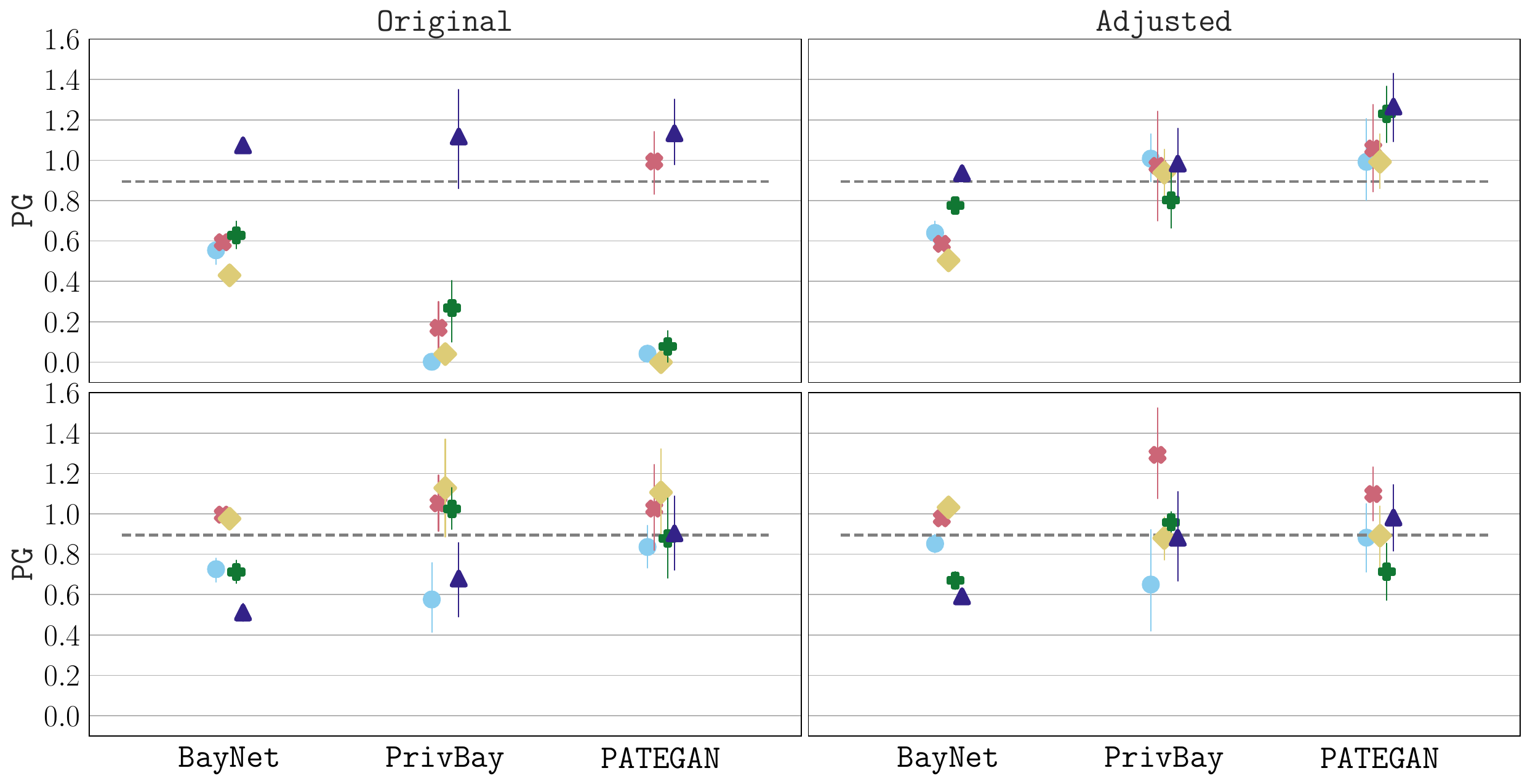}
	\caption{Per-record privacy gain for five outlier targets records from the \Texas~(\textit{top row}) and \Adult~(\textit{bottom row}) datasets under an attack using the $\Fhist$ feature set.}
	\label{fig:mia_dp}
\end{figure}

\parabf{Differential privacy violations.} \autoref{fig:mia_dp}~\emph{left} shows the results of this experiment for the \Texas\ (top row) and \Adult\ (bottom row) datasets under an attack using the histogram feature set $\Fhist$ -- the overall most effective attack. Both differentially private models were trained with privacy parameter $\varepsilon = 0.1$. Surprisingly, we find that neither the original implementation of $\PrivBay$ nor $\PATEGAN$ reliably prevents linkage attacks. Two out of the five outliers in the \Texas\ dataset achieve close to no gain (\tikzcircle{safeBlue} and \tikzdiamond{safeYellow} with $\PG < 0.1$). This low gain violates the theoretical lower bound on privacy provided by Yeom et al.~\cite{YeomGFJ18} (shown as a dashed line in \autoref{fig:mia_dp}). The bound given by Yeom et al.~\cite{YeomGFJ18} limits the expected advantage of the membership inference adversary to $\advantageMIA~\leq~e^\varepsilon~-~1$ which implies $\PG \geq 0.89$ for $\varepsilon = 0.1$.

\parabf{Unexpected leakage.} To understand these findings, we conducted an in-depth analysis of the design and implementation of both algorithms. Alongside some minor bugs, our analysis revealed that, while both models on paper fulfil their formal privacy definitions, their available implementations did not. Both $\PrivBay$ and $\PATEGAN$ require metadata about a model's training set to operate. For instance, data holders need to specify \textit{upfront} the range of numerical attributes and the possible values of categorical attributes. To improve usability, the existing implementations of $\PrivBay$ and $\PATEGAN$ learn this metadata \emph{directly} from the input dataset. This process, in which algorithmic decisions are based on the raw data, violates important assumptions of the differential privacy model~\cite{DworkA14}. 
This discovery explained our previous results: Both models extracted crucial information from their input data in a process \emph{not covered by the formal privacy guarantee}. As a result, targets with rare categorical attributes or whose presence affects the ranges of numerical attributes remained highly vulnerable to our attack. Their presence in a model's training set became detectable due to the occurrence of new categories or a shift in the ranges of continuous attributes in synthetic datasets sampled from the trained model.

To avoid this leakage, we patched the $\PrivBay$ and $\PATEGAN$ implementations so that both models obtain metadata as an \textit{independent} input to their training process. In our experiments with the \Texas\ dataset, we used the publicly available data description to define possible categorical values and a disjoint subset of the data (population records from a different year) to obtain an estimate of the expected ranges of numerical attributes. For the \Adult\ dataset, where no comparable metadata is available, we used the dataset to estimate categories and ranges and generalised each range to hide the exact value of outlier targets.

\autoref{fig:mia_dp}~\emph{right} shows the results of our evaluation under the patched implementations. For most outliers, privacy gain is now bounded by its differential privacy guarantee. 
For those targets where the expected gain remains below its bound, the remaining gap can likely be explained either by other aspects of the model's implementation that violate theoretical assumptions and we were not able to find in our analysis, or due to correlations between the datasets used to derive the necessary metadata and the model's training set.
Further work is needed to fully understand this problem.

\parabf{Conclusions.} Differentially private generative models can provide a significantly higher privacy gain with respect to linkage attacks than traditional data synthesis algorithms. To achieve the desired protection it is necessary that, besides a theoretically sound design, the models' implementation and operational environment does not break any of the privacy definition's theoretical assumptions. Our evaluation confirms that otherwise there is no guarantee that outliers will be protected from linkage attacks.

While in our experimental setup we were (mostly) able to avoid undesired privacy leakage through metadata, it is unlikely that in practice data holders will be able to follow our example. Data holders likely do not have access to either a disjoint subset or a public dataset from the same distribution that would allow them to define metadata that fits the raw data they would like to share. Synthetic data sharing is often motivated by the \emph{unique value} of sensitive dataset that are limited in size. This implies that, in practice, data holders might struggle to achieve the desired strict privacy guarantees, or face a large utility loss when either using public data or splitting the available data to derive the necessary metadata. In \autoref{subsec:put_utility}, we empirically demonstrate this tradeoff.

\section{Does synthetic data improve the privacy-utility tradeoff of sanitisation?}
Synthetic data is often presented as ``a new, better alternative to sanitised data release [...] that \emph{not only maintains the nuances of the original data}, but does so without endangering important pieces of personal information''~\cite{BellovinDR19}. Our results in previous sections show that synthetic data fulfils the latter part of this promise only partially. Data synthesis algorithms without any formal privacy guarantees leave outliers vulnerable to linkage attacks. Differentially private generative models, although hard to implement, reduce these risks. This makes differentially private synthetic data generation look like a promising alternative to traditional sanitisation. The question remains, however, whether synthetic data can achieve a higher gain in privacy \emph{at a lower cost in utility} compared to traditional sanitisation.

In this section, we assess the privacy-utility tradeoff of synthetic data publishing and compare it to that of traditional sanitisation. For our comparison, we implement a sanitisation procedure described by NHS England~\cite{SynAE20} and assess its privacy gain with respect to the risk of linkability (see \autoref{subsec:mia_game}) and the risk of inference formalised in \autoref{subsec:put_attribute}.

\parabf{NHS Sanitisation procedure.} A sanitisation procedure $S \gets \San(R)$ is a deterministic function that applies a set of pre-defined row-level transformations to the input data $R$ to produce a sanitised dataset $S$ that fulfils a heuristic privacy definition~\cite{Sweeney02}. Common transformations are generalisation, perturbation, or deletion of single rows~\cite{Art29WP}.
Following the details given in~\cite{SynAE20}, we implemented a simple sanitisation procedure $\San$ that reduces the granularity of categorical attributes through grouping, generalises any granular timing or geographical information, removes any rows with rare categorical values, caps numerical values to the attribute's $95\%$ quantile, and enforces $k$-anonymity for a pre-defined set of demographic attributes.

\subsection{Privacy gain with respect to linkability}\label{subsec:put_mia}

To compare the privacy gain of sanitised and synthetic data publishing, we repeat the experimental procedure from \autoref{subsec:mia_results}. We adapt the game so that the challenger $\chGame$, instead of generating a synthetic dataset $S$ from a trained model $\GMtrained{R}$ (lines 10 to 11 in \autoref{fig:mia_game}), produces a sanitised version of $R$ through a pre-defined sanitisation procedure $S \gets \San(R)$. When the adversary receives a sanitised dataset ($\public = 1$), she first attempts literal record linkage. Only if the adversary can not uniquely identify a record that matches the target, she attempts classification. As in previous sections, we first train the adversary on a reference dataset $R_\adv$ and then instantiate the game multiple times for each of the selected targets.

\begin{figure}
	\includegraphics{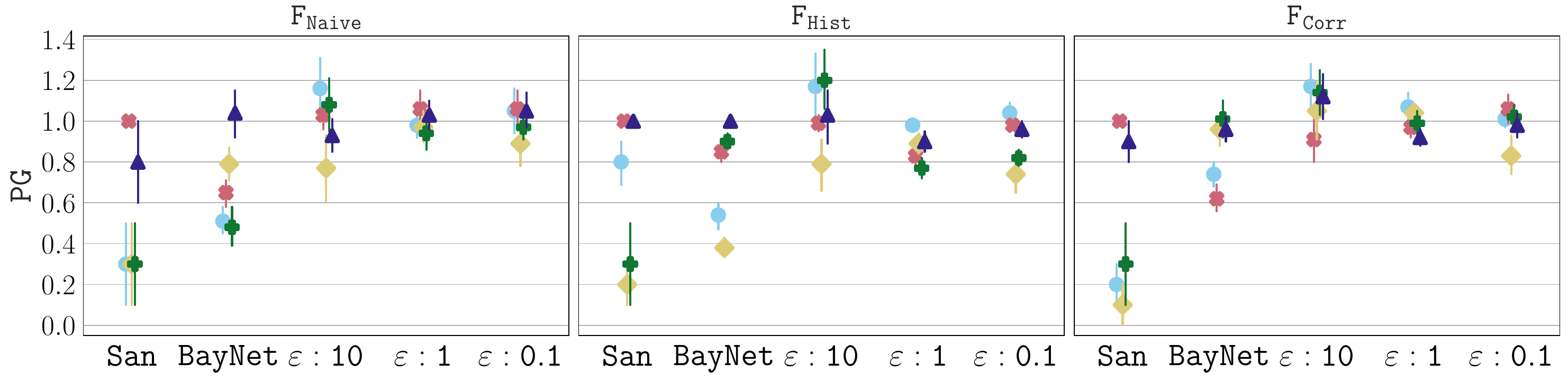}
	\caption{Per-record privacy gain for five outlier target records from the \Texas\ dataset under three different attacks using three distinct feature sets.}
	\label{fig:mia_san}
\end{figure}

\autoref{fig:mia_san} compares the results of this experiment for the five outlier targets from the \Texas\ dataset for three different data sharing mechanisms: traditional sanitisation $\San$ with $k=10$, synthetic data produced by $\BayNet$-trained models, and differentially private synthetic data sampled from $\PrivBay$ models with varying $\varepsilon$ values. The same experiment on the \Adult\ dataset yields similar results.

As expected, the privacy gain of row-level sanitisation tends to be binary: Target records that are likely to be removed from the shared dataset receive close to perfect gain (\tikztriangle{safePurple} and \tikzcrossrot{safePink} with $\PG \geq 0.8$ under all three feature sets). Others remain highly vulnerable to linkage attacks and receive a substantially lower gain (\tikzcircle{safeBlue},\tikzdiamond{safeYellow}, \tikzcross{safeGreen} with $\PG \leq 0.3$ for at least one attack). 

$\BayNet$ improves the privacy gain for the latter group: The three targets that under sanitisation receive close to no protection from linkage attacks using the naive feature set $\Fnaive$ obtain a higher minimum gain (\tikzcircle{safeBlue}, \tikzcross{safeGreen}, \tikzdiamond{safeYellow} with $\PG \geq 0.48$). Differentially private model training further improves protection and minimum gain increases as $\varepsilon$ decreases ($\PG = 0.77$ for target \tikzdiamond{safeYellow} under $\varepsilon = 10$ and $\PG = 0.97$ under $\varepsilon = 1.0$). This indicates that synthetic data produced by either model ($\BayNet$ and $\PrivBay$) hides changes in the raw data features caused by the target's presence and prevents the adversary from inferring the target's secret.    
This gain in privacy, however, is not constant across the population. One out of the five targets actually loses protection from linkage attacks when sharing a synthetic instead of the sanitised dataset (\tikzcrossrot{safePink} with $\PG = 0.62$ for $\BayNet$ and $\PG = 0.91$ for $\PrivBay$ with $\varepsilon = 10$ instead of $\PG = 1.0$ for $\San$ under $\Fcorr$).

This variability in privacy gain highlights one of the major drawbacks of synthetic data sharing as a privacy mechanism: unpredictability. Due to the deterministic nature of row-level sanitisation, the privacy gain of traditional anonymisation is largely predictable. The high gain in privacy for \tikzcrossrot{safePink} and \tikztriangle{safePurple} under $\San$ is constant across all three feature sets. In contrast, the privacy gain under $\BayNet$ and $\PrivBay$ is much more variable. Before model fitting and an empirical analysis, it is not possible to predict whether an individual record's signal will be preserved and what will be its minimum privacy gain.

\vspace*{-2mm}
\subsection{Privacy gain with respect to attribute inference}\label{subsec:put_attribute}
The risk of linkability is not the only concern in the context of privacy-preserving data sharing (see \autoref{sec:framework}). Data anonymisation also aims to protect individuals in the raw data from inference attacks. The risk of inference describes the concern that an adversary might ``deduce, with significant probability, the value of an attribute from the values of a set of other attributes''~\cite{Art29WP}. 

\vspace*{-2mm}
\subsubsection{Formalising attribute inference}
To evaluate privacy gain with respect to attribute inference, we define a privacy game similar to the attribute inference experiment proposed by Yeom et al.~\cite{YeomGFJ18}.

\begin{figure}[h]
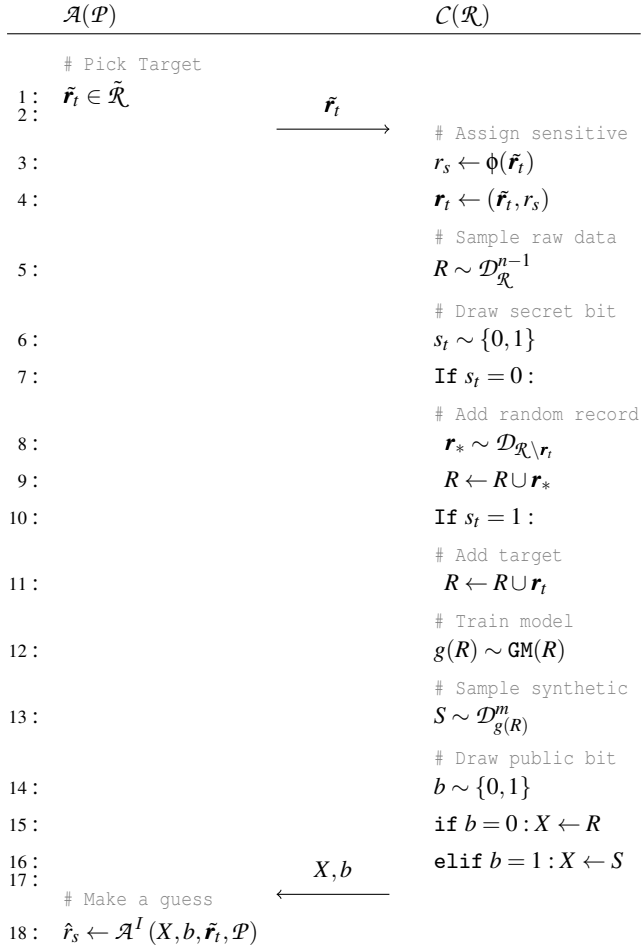

\newcommand{\comment}[1]{\textcolor{gray}{\footnotesize{\texttt{\# #1}}}}
 \pseudocodeblock[colsep=0em,skipfirstln, linenumbering]{
 	\pcskipln
 	\advGame(\prior) \< \< \chGame(\pop) \\[0.1\baselineskip][\hline]\pcskipln
 	\< \< \\[-0.5\baselineskip]
	\comment{Pick Target}\< \<\\[-0.2\baselineskip]
 	\targetpartial \in \tilde{\pop}\< \< \\[-1\baselineskip] \pcskipln
 	\< \sendmessageright*[1.5cm]{\targetpartial} \<\\[-1\baselineskip]
	\< \<\comment{Assign sensitive}\\[-0.2\baselineskip]
 	\< \< \sensitive \gets \phi(\targetpartial)\\ \pcskipln
	\< \< \target \gets (\targetpartial, \sensitive)\\
	\< \<\comment{Sample raw data}\\[-0.2\baselineskip] \pcskipln
	\< \< R \sim \distPop^{\sizeR-1}\\ 
	\< \<\comment{Draw secret bit}\\[-0.2\baselineskip]
	\< \< \secret \sim\{0,1\}\\ \pcskipln
 	\< \< \mathtt{If}~\secret=0:\\ 
	\< \<\comment{Add random record}\\[-0.2\baselineskip]
 	\< \< \;\;\record_* \sim \mathcal{D}_{\pop \setminus \target}\\
 	\< \< \;\;R \gets R \cup \record_*\\\pcskipln
 	\< \< \mathtt{If}~\secret=1:\\
	\< \<\comment{Add target}\\[-0.2\baselineskip]\pcskipln
 	\< \< \;\;R \gets R \cup \target\\
	\< \<\comment{Train model}\\[-0.2\baselineskip]\pcskipln
 	\< \< \GMtrained{R} \sim \trainGM{R}\\
	\< \<\comment{Sample synthetic}\\[-0.2\baselineskip]\pcskipln
 	\< \< S \sim \distGM{R}^\sizeS\\
	\< \<\comment{Draw public bit}\\[-0.2\baselineskip]
 	\< \< \public\sim\{0,1\}\\
	\< \< \mathtt{if}~\public = 0: X \gets R\\
	\< \< \mathtt{elif}~\public = 1: X \gets S\\[-1\baselineskip]\pcskipln
 	\< \sendmessageleft*[1.5cm]{X, b} \<\\[-1\baselineskip]
	\comment{Make a guess}\< \<\\[-0.2\baselineskip]
 	\sensitivehat \gets \advAI\left( X, \public,  \targetpartial, \prior \right)\< \<
 	}
\caption{Attribute inference privacy game}\label{fig:attribute_game}
\end{figure}

In the attribute inference game, shown in \autoref{fig:attribute_game}, the adversary only has access to a \emph{partial} target record $\targetpartial = (r_1, \cdots, r_{\nAttr-1})$ and aims to infer the value of a sensitive, unknown attribute $\sensitive$. At the start of the game, the adversary picks a target from the population $\tilde{\pop}$, a set of records from the same domain as $\pop$ but with the sensitive attribute removed. The challenger receives the partial target record and assigns it a secret value $\sensitive \gets \phi(\targetpartial)$ where $\phi$ represents the projection of a partial record from $\tilde{\pop}$ into the domain of the sensitive attribute according to the distribution $\distPop$. $\chGame$ then follows the same procedure as in the linkability game. The adversary obtains the dataset $X$ and the public bit $\public$ and outputs a guess about the target's sensitive attribute value $\sensitivehat$. 
This game can be easily adapted for sanitisation by replacing lines 12 and 13 to produce a sanitised version of $R$ through a pre-defined sanitisation procedure, $S \gets \San(R)$.

Similar to \autoref{eq:mia_adv}, we define the adversary's advantage to assess the leakage of publishing dataset $X$ with respect to attribute inference as:

\begin{equation}\label{eq:ai_adv}
	\advantageAI(X, \targetpartial) \triangleq \Prob{\sensitivehat = \sensitive | \secret = 1} - \Prob{\sensitivehat = \sensitive| \secret = 0}
\end{equation}

where $\sensitivehat = \advAI\left( X, \public, \targetpartial, \prior \right)$ is the adversary's guess about the target's sensitive attribute $\sensitive$ given dataset $X$ and prior knowledge $\prior$. 

\parabf{Adversarial strategy.} The procedure to estimate $\advantageAI$ and the adversary's strategy to make a guess about the target's sensitive value depends on the domain of the sensitive attribute $\sensitive$, the value of the public bit $b$ and whether $S$ is a synthetic or sanitised dataset.

When $\chGame$ publishes a raw or sanitised dataset, the adversary first attempts to infer the missing value via record linkage~\cite{MachanavajjhalaGKV06, ReiterM09, DrechslerR10}. If the adversary can link the target to a unique record in the dataset $X$ based on its known attributes, she can reconstruct the target's missing value with probability $\Prob{ \advAI\left( X, \public, \targetpartial, \prior \right) = \sensitive | \secret = 1} = 1$. 

When linkage fails, i.e., $\chGame$ publishes a raw dataset without the target, a sanitised dataset that hides the target's presence or a synthetic dataset, the adversary uses the published data to \emph{train a supervised ML model to predict the target's sensitive value} based on the known attributes $\targetpartial$. 
To learn a mapping from known to sensitive attributes, the adversary splits the dataset $X$ into two parts: A feature matrix $\tilde{X}$ that contains the values for all attributes known to the adversary and a vector $\bm{x}_s$ with the corresponding sensitive attribute values. Depending on the domain of $\bm{x}_s$, the adversary can either train a regression or classification model using $\tilde{X}$ as input features and $\mathbf{x}_s$ as labels. The trained attack model, denoted as $\PMtrained{\cdot}$, takes as input a partial record containing the set of known attributes and outputs a guess about the label $\hat{x}_s \gets h(\tilde{\mathbf{x}})$.

\parabf{Implementation.} We implement the adversary $\advAI$ as an instantiation of our framework's $\PrivacyAttack$ class (see Appendix~\ref{ap:implementation}).
For continuous sensitive attributes with $\sensitive \in \mathbb{R}$, we implement the attack $\PMtrained{\cdot}$ using a simple linear regression model provided by the \texttt{sklearn} library~\cite{PedregosaVGMTGBPWDVPCBPD11}. We centre all features extracted from the input data and fit a linear model without intercept. The model fits linear coefficients that minimise the root mean squared error between the observed and predicted target values. We analytically calculate the adversary's probability of success $\Prob{\sensitivehat = \sensitive | \secret}$ as the likelihood of the true value under the learned linear coefficients (see Appendix \ref{ap:attribute} for details). For categorical attributes, we use a simple Random Forests classifier as attack model and estimate the attack's success via its classification accuracy.
  
\begin{figure}
	\includegraphics{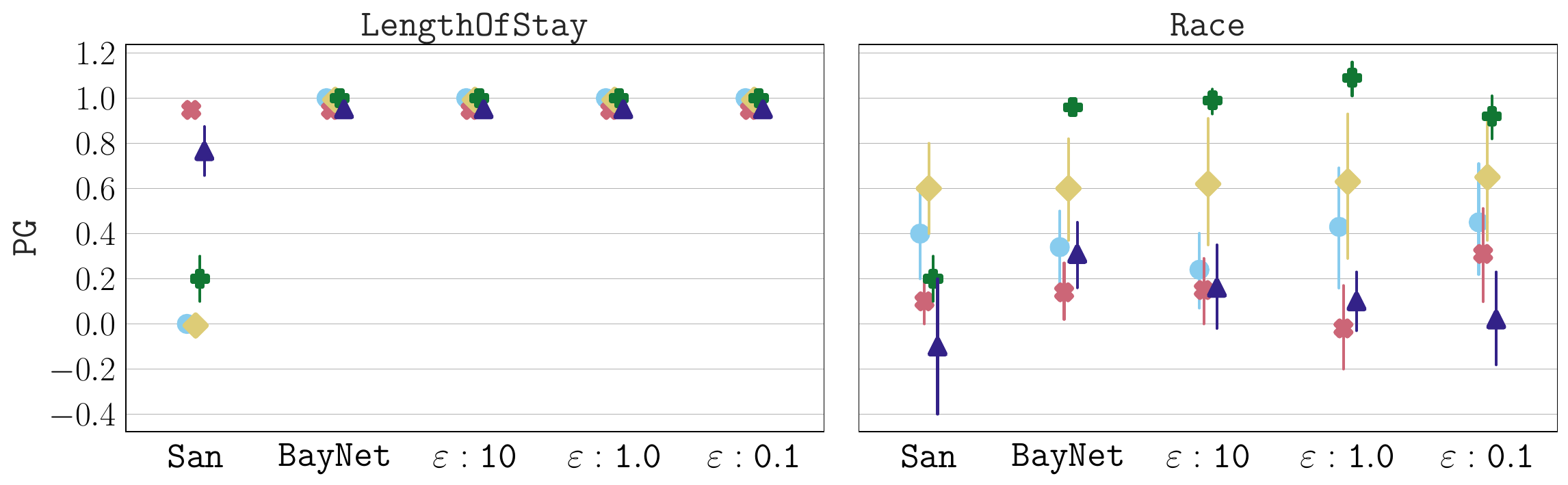}
	\caption{Per-record privacy gain for five outlier target records from the \Texas\ dataset for two distinct sensitive attributes.}
	\label{fig:attribute_gain}
\end{figure}

\parabf{Empirical evaluation.} \autoref{fig:attribute_gain} shows the privacy gain of the three data release mechanisms, sanitisation via $\San$ and synthetic data produced by $\BayNet$ and $\PrivBay$-models with varying $\varepsilon$ values, for the five outlier targets from the \Texas\ dataset for two distinct sensitive attributes. We chose one continuous (\texttt{LengthOfStay}) and one categorical attribute (\texttt{Race}) that might be considered sensitive patient information. 

For the continuous attribute \texttt{LengthOfStay}, synthetic data produced by either $\BayNet$ or $\PrivBay$ provides close to perfect gain for all five targets while row-level sanitisation via $\San$ marginally reduces the adversary's advantage for three out of the five targets (\tikzcircle{safeBlue}, \tikzcross{safeGreen}, \tikzdiamond{safeYellow} with $\PG \leq 0.2$).
This implies that, in contrast to the sanitised datasets, synthetic data does not preserve the targets' signal. Even when the generative model's training set includes the target record, synthetic data sampled from the trained model does not contain any patterns that allow the adversary to infer the target's sensitive value and hence $\advantageAI(S, \target) \ll \advantageAI(R, \target)$. 
When the attack targets the categorical attribute \texttt{Race}, the privacy gain of synthetic data publishing does not significantly increase over that of sanitised data publishing. Even differentially private generative model training with $\varepsilon = 0.1$ does not guarantee a high privacy gain.

\begin{figure}
	\includegraphics{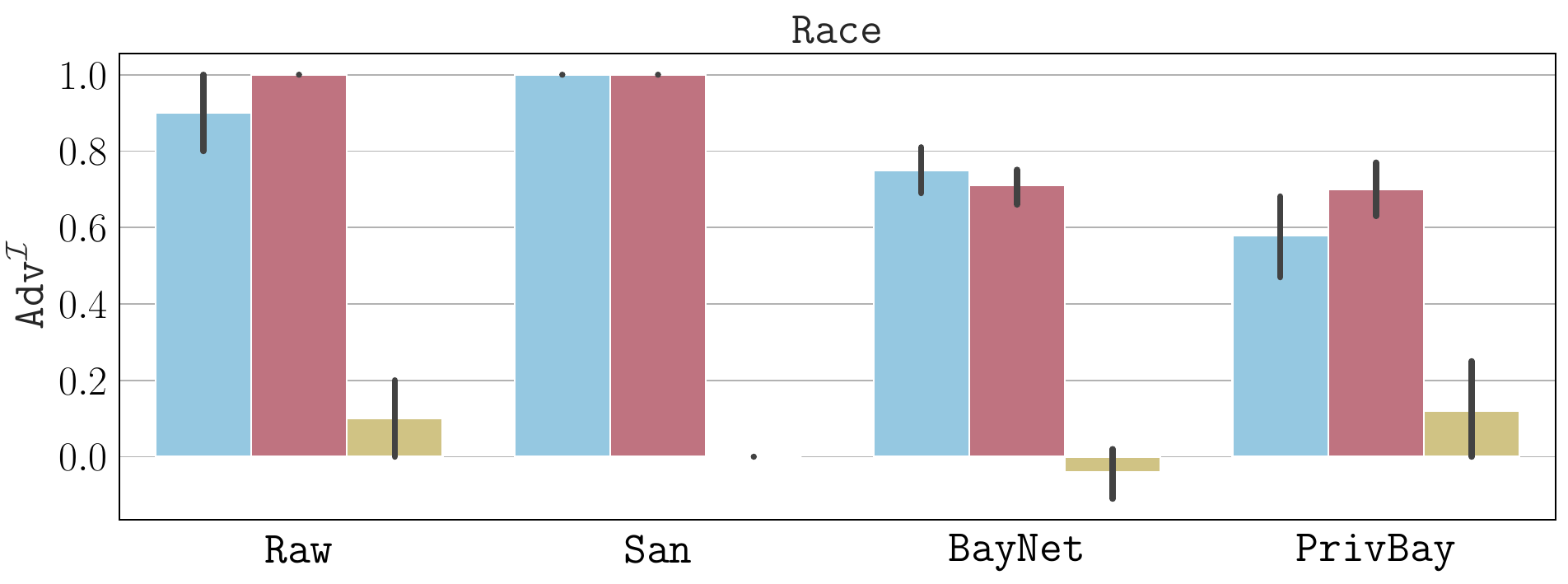}
	\caption{Probability of success and advantage for an attribute inference attack on attribute \texttt{Race} on the \Texas\ dataset. \protect\tikzbar{safeBlue}~$\Prob{\sensitivehat = \sensitive | \secret = 0}$, \protect\tikzbar{safePink}~$\Prob{\sensitivehat = \sensitive | \secret = 1}$, and \protect\tikzbar{safeYellow}~$\advantageAI$. Data shown for $\PrivBay$ with $\varepsilon = 1$.}\label{fig:attribute_adv}
\end{figure}

To explain this low gain, we plot in \autoref{fig:attribute_adv} for the target marked as \tikzcrossrot{safePink} the adversary's probability of success on dataset $X$ when the target is in the dataset (\tikzbar{safePink} $\secret = 1$), when it is not (\tikzbar{safeBlue} $\secret = 0$), and the resulting adversary's advantage (\tikzbar{safeYellow} $\advantageAI$). The figure shows that the low privacy gain observed in \autoref{fig:attribute_gain} results from the fact that the adversary's advantage is already small when the adversary receives the raw data $R$. Thus, publishing a sanitised or synthetic data instead of the raw data does not lead to any substantial gain in privacy for this target.    

\parabf{Conclusions.} Depending on the attribute targeted by the attack, the privacy gain of synthetic and sanitised data publishing varies substantially. A low gain in privacy either indicates that the anonymised data still contains enough information specific to the target record to give the adversary a significant advantage or it implies that already publishing the raw data did not incur a significant privacy loss. Where necessary, our framework allows data holders to make a distinction between those two cases based on the reported privacy loss. In both cases, however, a low gain signals that publishing the synthetic or sanitised data does not provide any improvement over publishing the raw data $R$.
We also observe that there are records for which privacy gain is high: the dataset published in place of the raw data successfully hides information about these records. While this might be good news for privacy, it comes, as we show in the next sections, at a cost in utility.

\subsection{Utility loss comparison}\label{subsec:put_utility}
The promise of synthetic data is that its improvement in privacy gain over traditional sanitisation comes at a negligible cost in utility. In this section, we empirically evaluate the utility loss of synthetic data sharing and compare it to that of row-level sanitisation. 

\parabf{Utility metrics.} Besides its potential privacy benefits, sharing a synthetic in-place of the original dataset incurs certain risks, such as the risk of false conclusions~\cite{ArnoldN20} or the risk of exacerbating existing biases in the data~\cite{ChengSD21}.
The goal of our utility evaluation is to assess to which extent the privacy gain of synthetic data observed in previous sections (see \autoref{subsec:put_mia} and \ref{subsec:put_attribute}) increases these risks and reduces data utility.

The concepts of utility and utility loss of course are highly dependent on the data use case and different utility metrics might yield vastly different results~\cite{XuSCV19, RosenblattLPL20}. Therefore, in practice data holders should conduct their own evaluation based on appropriate utility definitions when weighing off the risks and benefits of (anonymised) data sharing.

In this work, we chose a set of simple utility function that aim to cover a wide range of synthetic data use cases suggested in the literature and reported to us by practitioners.
First, we evaluate in~\autoref{subsec:average_utility} the utility of synthetic data for use cases that rely on aggregate population metrics, such as reporting of summary statistics or training machine learning models~\cite{RosenblattLPL20, SDGym}.
Second, in \autoref{subsec:outlier_utility}, we turn to one of the main selling points brought forward by proponents of synthetic data as a privacy technology: That it enables the analysis of more fine-grained statistical patterns than aggregate query release mechanisms, \textit{including the analysis of outliers}. Financial fraud and medical anomaly detection are two of the most commonly suggested synthetic data use cases largely based on the analysis of outliers~\cite{TuckerWRM20, MostlyAI_Fraud}.

The latter class of use cases further motivates us to focus on the privacy gain and utility loss of outlier records. A simple way to improve privacy gain for these most vulnerable records would be to remove them from the dataset, as the high privacy gain for targets \tikzcross{safePink} and \tikztriangle{safePurple} under $\San$ in \autoref{fig:mia_san} demonstrates.
Directly removing vulnerable records from the raw data increases privacy gain but severely impacts the data's utility, e.g. in healthcare ``this can sometimes mean missing out on important data that could be used to help future patients''~\cite{TuckerWRM20}. For this reason, we assess in \autoref{subsec:outlier_utility} whether synthetic data leads to a high privacy gain for outlier records while preserving their utility benefits for similar records.

\subsubsection{Average utility loss}\label{subsec:average_utility}
\parabf{Machine learning utility.} We first measure the utility loss of publishing a synthetic or sanitised dataset $S$ in place of the raw data $R$ as the decrease in \emph{average prediction accuracy over a hold out set} for a prediction model trained on $S$ instead of $R$. Due to the limited size of the \Adult\ dataset, we focus our evaluation on the \Texas\ dataset. 
We created a pre-processed dataset that contained all publicly available inpatient records for the years of $2013$ and $2014$. We denote records from $2013$ as the train population $\pop_{Train}$, and records from $2014$ as the test set $\pop_{Test}$. In each experiment, we sample a raw dataset of size $\sizeR$ from $\pop_{Train}$ and use it as training set for a prediction model $\PMtrainedX{R}{\cdot} \sim \trainPM{R_{Train}}$. We then train a generative model $\GMtrained{R_{Train}}$ on the \emph{same raw dataset} and sample multiple copies of synthetic data $S_i$ from this generative model. We use these synthetic datasets to train classifiers $\PMtrainedX{S}{\cdot}$. We follow an analogous procedure for sanitisation, where we sanitise $R_{Train}$ according to the NHS sanitisation procedure before training the classifier. Finally, we evaluate the accuracy of each of the trained classifiers $\PMtrainedX{X}{\cdot}$ on a test set sampled from $\pop_{Test}$.

\begin{figure}
	\includegraphics{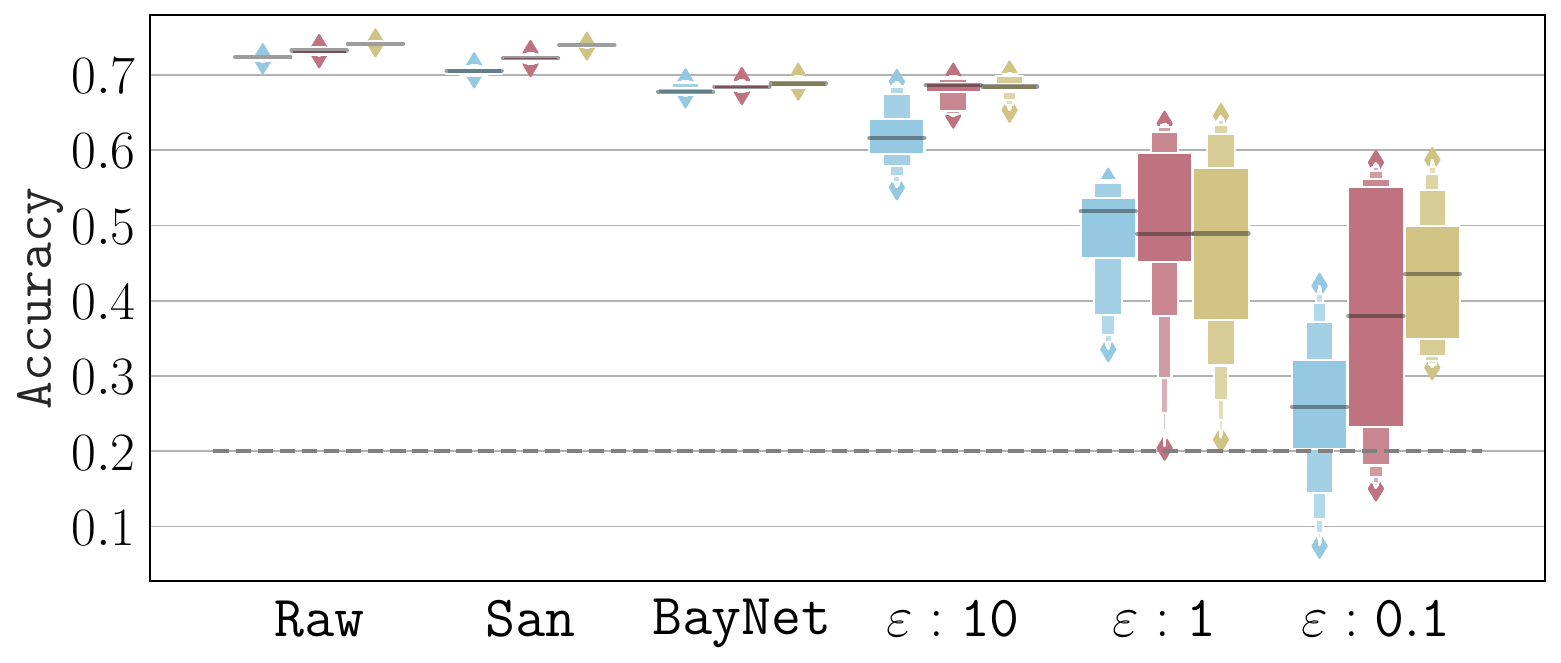}
	\caption{Test accuracy of prediction models for attribute \texttt{RiskMortality} in the \Texas\ dataset for three dataset sizes (\protect\tikzbar{safeBlue}~$n=1000$, \protect\tikzbar{safePink}~$n=2000$ and \protect\tikzbar{safeYellow}~$n=5000$). The grey dotted line shows the random guess baseline.}
	\label{fig:util_avg}
\end{figure}

\autoref{fig:util_avg} shows the results of this experiment for a multi-classification task on attribute \texttt{RiskMortality} with $5$ classes for three different training set sizes (\tikzbar{safeBlue}~$n=1000$, \tikzbar{safePink}~$n=2000$, and \tikzbar{safeYellow}~$n=5000$). \autoref{fig:mia_sizes} in the Appendix \ref{ap:utility} shows the corresponding privacy gain for various dataset sizes.
The random baseline guess rate is indicated with a dotted grey line.
With \tikzbar{safeBlue}~$n=1000$, classifiers trained on the raw data $\PMtrainedX{R}{\cdot}$ achieve an average test accuracy of $72.2\%$ which is comparable to models trained on the sanitised dataset with $70.5\%$. Using a synthetic version of the data as training set leads to a significant utility loss: The classifier's average test accuracy drops to $68.0\%$ when trained on synthetic data produced by $\BayNet$. Differentially private model training further widens this gap. Even with privacy parameter values as high as $\varepsilon = 10$ the classifier's mean accuracy with $62.0\%$ remains $10$ points below that of a model trained on the raw data.
Increasing the model's training set size has a slight effect on utility. While classifiers trained on the raw data achieve an average accuracy of $74.2\%$ for \tikzbar{safeYellow} $n=5000$, average accuracy under synthetic data produced by $\BayNet$ and $\PrivBay$ with $\varepsilon = 10.0$ increases to $68.9\%$ and $68.4\%$, respectively. 

\begin{figure}
	\includegraphics{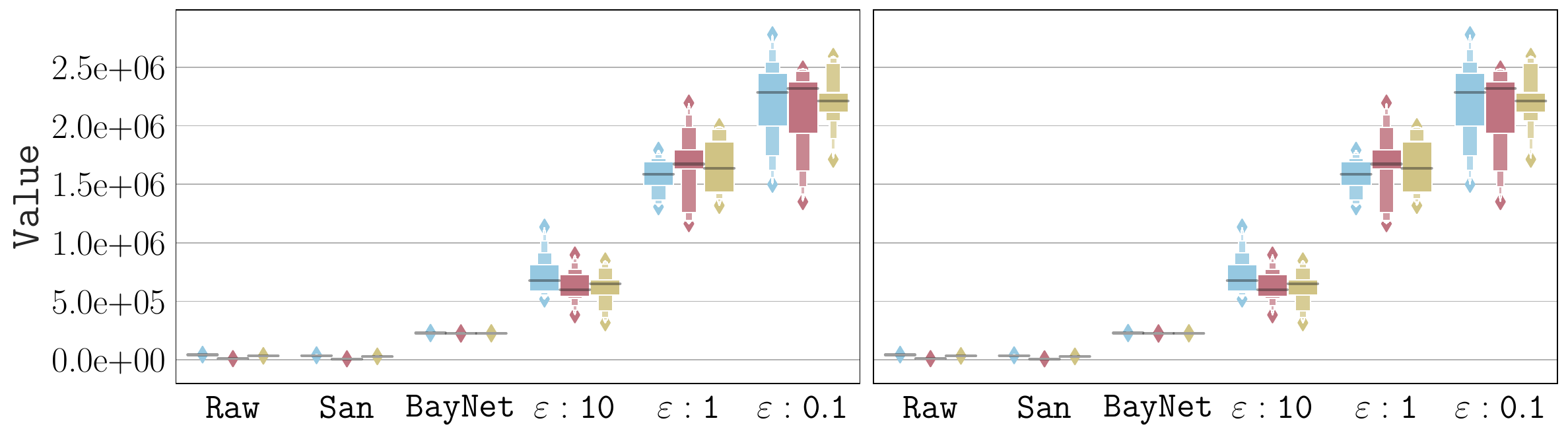}
	\caption{Mean (\textit{left}) and median (\textit{right}) for attributes \protect\tikzbar{safeBlue}~\texttt{TotalCharges}, \protect\tikzbar{safePink}~\texttt{TotalChargesAccomm}, and \protect\tikzbar{safeYellow}~\texttt{TotalChargesAncil} from the \Texas~dataset.}\label{fig:utility_stats}
\end{figure}

\parabf{Summary statistics.} We use the discrepancy of the mean (\autoref{fig:utility_stats} \textit{left}) and median (\textit{right}) of three continuous attributes between raw, sanitised, and synthetic datasets as additional utility loss measures. These simple summary statistics are common metrics used for reporting. While data sanitisation largely preserves the raw data's statistics, synthetic datasets sampled from all models significantly differ from the raw data. For instance, the empirical mean of attribute \tikzbar{safePink}~\texttt{TotalChargesAccomm} in raw datasets $R$ sampled from the population $\distPop$ ranges between $9$K and $11$K. In contrast, synthetic data sampled from $\BayNet$ models trained on the raw data produces values ranging from $218$K to $232$K. The deviation between the raw and synthetic datasets' characteristics further increases with decreasing privacy parameter $\varepsilon$ and grows to an error of \emph{multiple orders of magnitude}.

\begin{figure}
	\includegraphics{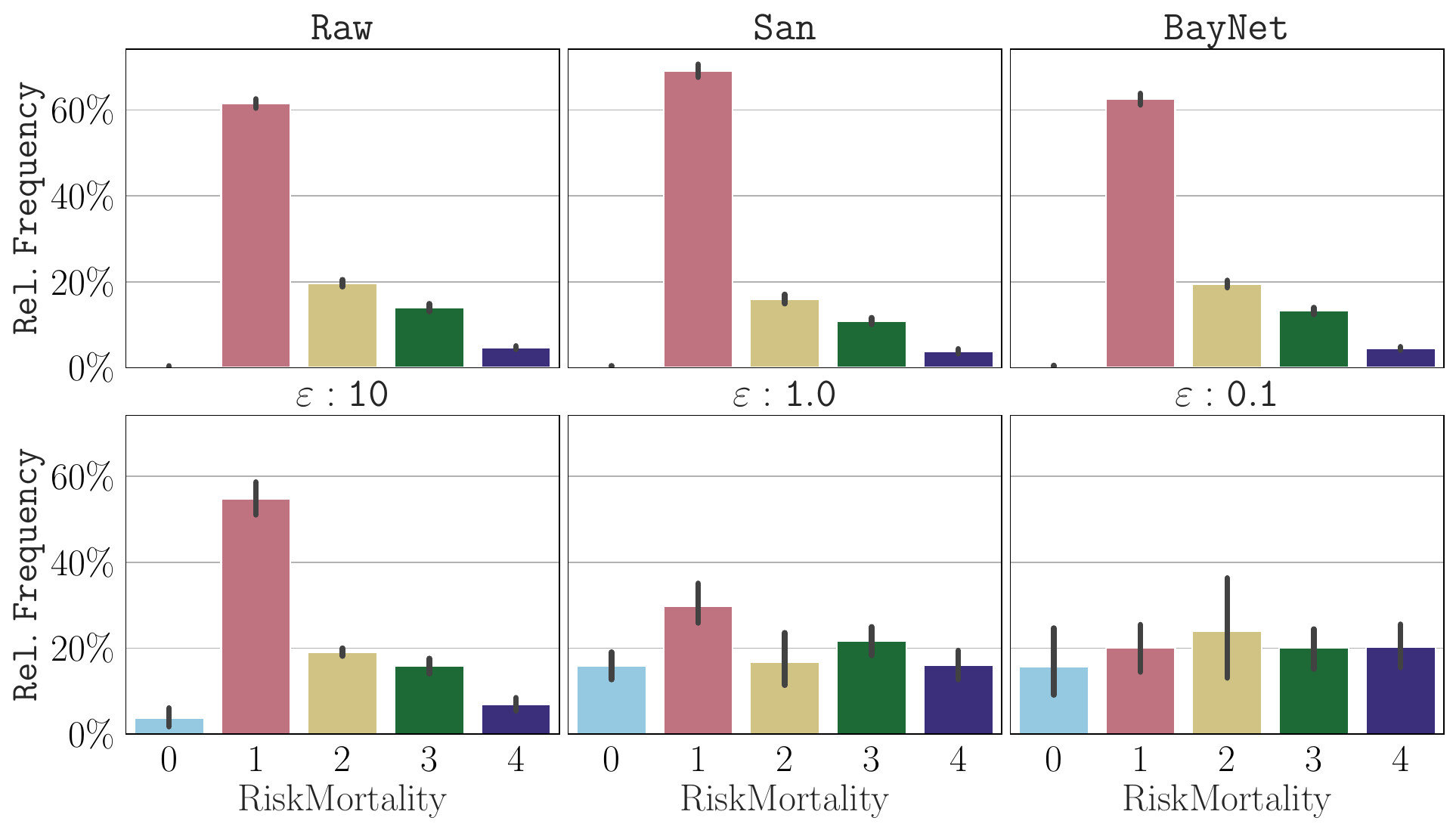}
	\caption{Marginal frequency counts for attribute \texttt{RiskMortality} from the \Texas~dataset.}
	\label{fig:utility_marginals}
\end{figure}

We observe the same trend on the marginal frequency counts over the categorical attribute \texttt{RiskMortality}, another common reporting task (see \autoref{fig:utility_marginals}). For instance, publishing a synthetic dataset sampled from a $\PrivBay$-trained model in place of the raw data would lead an analyst to overestimate the relative frequency of category $0$ by $3\%$ percentage points when $\varepsilon = 10$ and up to $15.3\%$ percentage points for $\varepsilon = 0.1$ with a growing variance of the mean error as $\varepsilon$ decreases.

These results highlight the risks of synthetic data sharing outlined in \autoref{sec:syntheticdef}. Generative models represent a \emph{lower-dimensional approximation} of the raw data's distribution and only capture a subset of the dataset's high-dimensional feature space. Synthetic data sampled from the trained model hence does not preserve all of the raw data's statistics and can lead to a large error on the derived insights. Synthetic data that accurately reflects the desired statistics would improve utility but not provide any privacy gain over directly publishing those aggregates.

\parabf{No free lunch.} In summary, the promising privacy gain of (differentially private) synthetic data publishing over traditional sanitisation shown in \autoref{fig:mia_san} comes at a significant cost in utility. Unsurprisingly, the higher a model's privacy gain, the higher its loss in utility. Even models with low theoretical privacy guarantees ($\varepsilon > 1$) may not provide the expected utility benefits and their utility loss may be prohibitive for many use cases.

The high utility loss of differentially private models can partially be explained by their reliance on metadata that needs to be derived independent of the model's training set (see \autoref{sec:dp}). This can lead to a \emph{significant difference} between the ranges in the raw data and the ones the data holder defines based on her background knowledge.
To optimise the utility of synthetic datasets, the metadata given as input to the generative model training must reflect the raw data characteristics as closely as possible. However, this increases the privacy leakage of the model and, as we show in \autoref{sec:dp}, in the most extreme case undermines its formal privacy guarantees. 

\begin{figure}
	\includegraphics{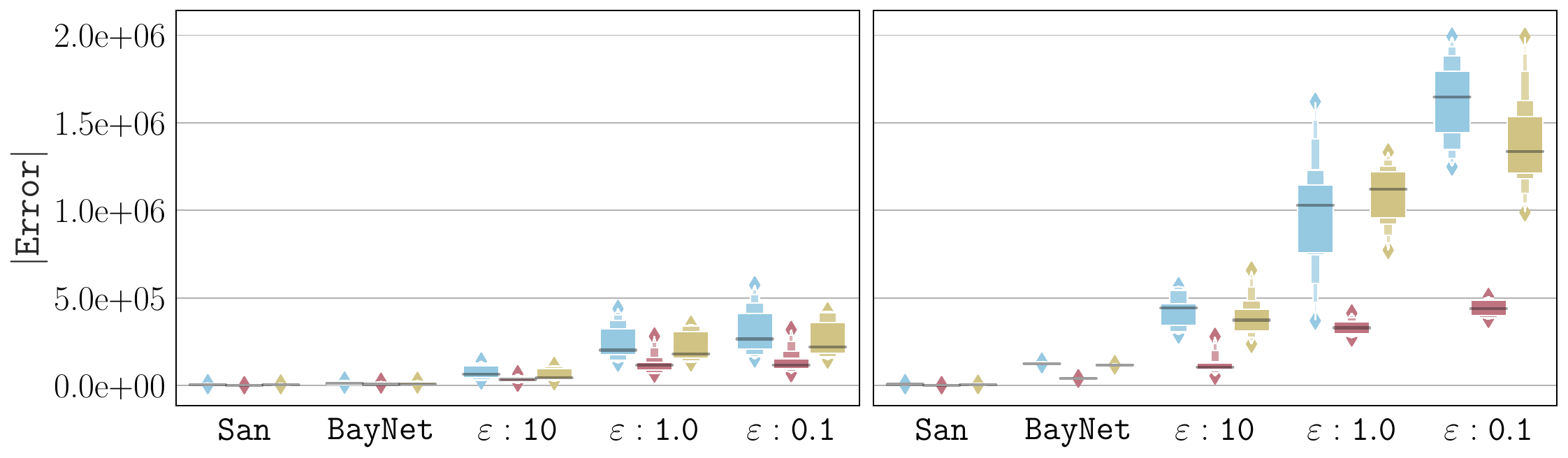}
	\caption{Absolute distance between the mean of attributes \protect\tikzbar{safeBlue}~\texttt{TotalCharges}, \protect\tikzbar{safePink}~\texttt{TotalChargesAccomm}, and \protect\tikzbar{safeYellow}~\texttt{TotalChargesAncil} in raw and synthetic datasets with metadata extracted from the raw data (\textit{left}) and metadata derived independently from the training set (\textit{right})}
	\label{fig:ranges_utility}
\end{figure}

\begin{figure}
	\includegraphics{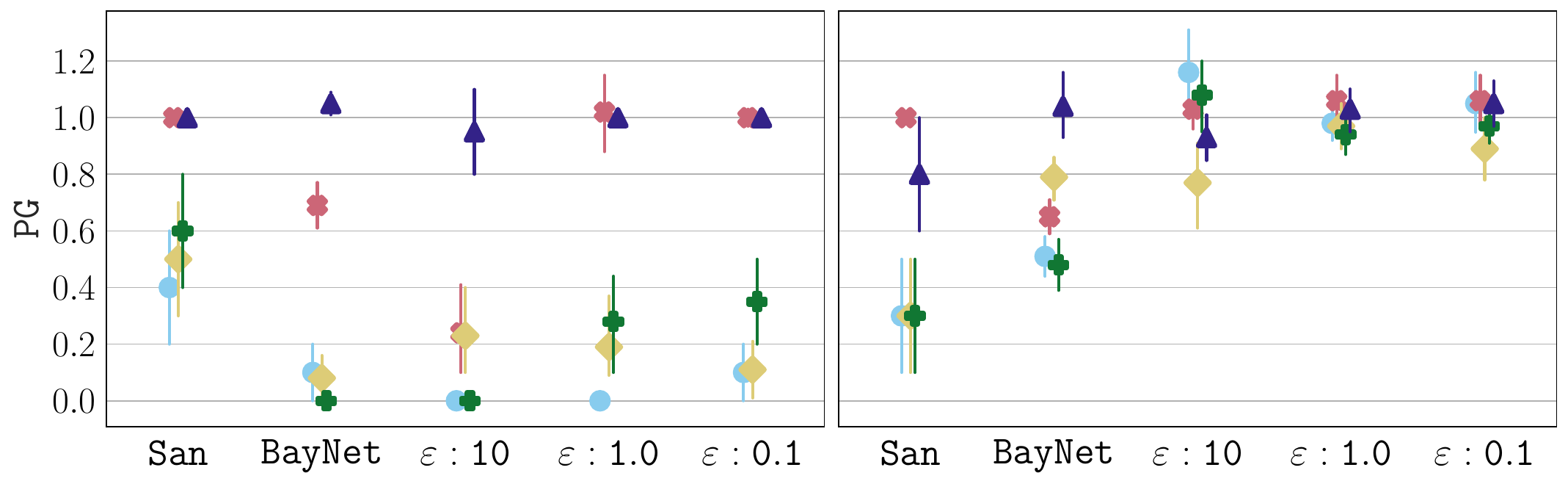}
	\caption{Per-record privacy gain for five outlier target records from the \Texas~dataset under an attack using the $\Fnaive$ feature set with metadata extracted from the raw data (\textit{left}) and metadata derived independently from the training set (\textit{right})}
	\label{fig:ranges_privacy}
\end{figure}

We demonstrate this tradeoff in \autoref{fig:ranges_utility} and \autoref{fig:ranges_privacy}.
\autoref{fig:ranges_utility} compares the absolute distance between the mean of three attributes in the raw data to the corresponding mean value in synthetic datasets sampled from models trained given different metadata as input. On the \textit{left}, all models were given the exact ranges of attributes in the raw data sample. On the \textit{right}, models used metadata learned from an independent population sample (see \autoref{sec:dp} for more details).
In the case where the model directly learns all metadata from its raw input data (\textit{left}), the utility loss of (differentially private) synthetic data is substantially reduced. For instance, the average distance between the mean of attribute \tikzbar{safePink}~\texttt{TotalChargesAccomm} in the raw and synthetic data shrinks from $215$K for a $\BayNet$ model trained with fixed ranges to $9$K for a model based on metadata tailored to its training set. The error of differentially private models similarly decreases (from $621$K to $34$K for $\PrivBay~\varepsilon: 10$) and with increasing $\varepsilon$ values converges to its non-private version. Despite this improvement the utility loss of all differentially private models we tested still remains far above that of non-private synthetic data generation or traditional sanitisation.
On the downside, as we show in \autoref{fig:ranges_privacy} (\textit{left} metadata tailored to training set, \textit{right} metadata derived from an independent population sample), the utility gain achieved through better metadata leads to a large loss in privacy. When the model's metadata is derived from the raw data, even models with low data utility provide close to no protection against linkage attacks ($\PG \leq 0.35$ for three out of five targets tested across all $\varepsilon$ values). Models that due to the use of metadata derived from an independent population sample result in a high privacy gain unfortunately do not provide sufficient data utility (see \autoref{fig:ranges_utility}). 

\subsubsection{Per-record utility loss}\label{subsec:outlier_utility}
In the previous section, we show that for use cases that aim to derive aggregate insights from sensitive datasets generative models \emph{with a high privacy gain} suffer from low data utility. However, average population metrics are not the main selling point of synthetic data. Synthetic data is often advertised as a solution for analysis tasks focused on more fine-grained statistical patterns that requires access to row-level data, in particular, the analysis of rare events and minority population subgroups. We hence now study whether synthetic data allows data holders to share datasets that \emph{retain the signal of outlier records} without risking their privacy.   

To study this relationship between privacy gain and utility loss, we slightly abuse the notion of advantage and formalise our utility metric as an advantage measure. This formalisation allows us to quantify the positive impact that the presence of a single target record in the training set has on a model's performance for individual test records. We define a utility game, shown in \autoref{fig:util_game}, played between an analyst $\advGame$ and a challenger $\chGame$. The analyst's goal is to train a predictive model $\PMtrained{\cdot}$ on a dataset $X$ published by the challenger that performs well on a chosen test record $\test$ from the population $\pop_{Test}$. The analyst chooses this test record and a target record $\target$ from the training population $\pop_{Train}$. The analyst sends $\target$ to the challenger, who follows the same procedure as in the linkability game and sends back the public bit $\public$ and the chosen dataset $X$. The analyst trains a model on dataset $X$ and uses it to predict the test record's label $\labelhat \gets \PMtrainedX{X}{\testpartial}$. The analyst wins the game if $\labeltest = \labelhat$.

\begin{figure}[h]
\newcommand{\comment}[1]{\textcolor{gray}{\footnotesize{\texttt{\# #1}}}}
 \pseudocodeblock[colsep=0em, skipfirstln, linenumbering]{
	\pcskipln
 	\advGame(\pop_{Test}) \< \< \chGame(\pop_{Train}) \\[0.1\baselineskip][\hline] \pcskipln
 	\< \< \\[-0.5\baselineskip]
	\comment{Pick target \& test}\< \<\\[-0.2\baselineskip] \pcskipln
 	\target \in \pop_{Train},~\test \in \pop_{Test}\< \< \\[-1\baselineskip]
 	\< \sendmessageright*[1.5cm]{\target} \<\\[-1\baselineskip]
	\< \<\comment{Sample raw}\\[-0.2\baselineskip]
	\< \< R \sim \distPop^{\sizeR-1}\\ \pcskipln
	\< \<\comment{Draw secret bit}\\[-0.2\baselineskip]
	\< \< \secret \sim\{0,1\}\\ \pcskipln
 	\< \< \mathtt{If}~\secret=0:\\
	\< \<\comment{Add random record}\\[-0.2\baselineskip]
 	\< \< \;\;\record_i \sim \mathcal{D}_{\pop \setminus \target}\\
 	\< \< \;\;R \gets R \cup \record_i\\ \pcskipln
 	\< \< \mathtt{If}~\secret=1:\\
	\< \<\comment{Add target}\\[-0.2\baselineskip] \pcskipln
 	\< \< \;\;R \gets R \cup \target\\
	\< \<\comment{Train model}\\[-0.2\baselineskip] \pcskipln
 	\< \< \GMtrained{R} \sim \trainGM{R}\\
	\< \<\comment{Sample synthetic}\\[-0.2\baselineskip] \pcskipln
 	\< \< S \sim \distGM{R}^\sizeS\\
	\< \<\comment{Draw public bit}\\[-0.2\baselineskip]
 	\< \< \public\sim\{0,1\}\\
	\< \< \mathtt{if}~\public = 0: X \gets R\\
	\< \< \mathtt{elif}~\public = 1: X \gets S\\[-1\baselineskip] \pcskipln
 	\< \sendmessageleft*[1.5cm]{X, \public} \<\\[-1\baselineskip]
	\comment{Train model}\< \<\\[-0.2\baselineskip]
 	\PMtrainedX{X}{\cdot} \sim \trainPM{X}\< \<\\
	\comment{Make guess}\< \<\\[-0.2\baselineskip] \pcskipln
	\labelhat \gets \PMtrainedX{X}{\testpartial}\\ \pcskipln
	}
	\caption{Utility game}\label{fig:util_game}
\end{figure}

Based on this game, we define the advantage that adding a record $\target$ to the dataset $R$ gives to record $\test$ as

\begin{equation}\label{eq:adv_util}
	\advantageU(X, \test, \target) \triangleq \Prob{\labelhat = \labeltest | \secret = 1} - \Prob{\labelhat = \labeltest | \secret = 0}
\end{equation}

where $\labelhat = \PMtrainedX{X}{\testpartial}$ is the label produced by a predictive model trained on data $X$ for test record $\test = (\testpartial, \labeltest)$ chosen by the analyst with feature set $\testpartial$ and label $\labeltest$; and $\secret$ indicates the presence of target record $\target$ in the raw dataset $R$. 

To empirically evaluate the utility loss of synthetic data publishing, we use the same split of the \Texas\ dataset as in the previous section. We ensured that the outlier targets used in previous experiments were included in the population $\pop_{Train}$ and re-used them as target records $\target$. We manually chose five test records in $\pop_{Test}$ that are semantically similar to those five target records. For instance, we selected records with rare categorical attribute values or continuous attributes outside the test population's $95\%$ quantile.

\begin{figure}
	\includegraphics{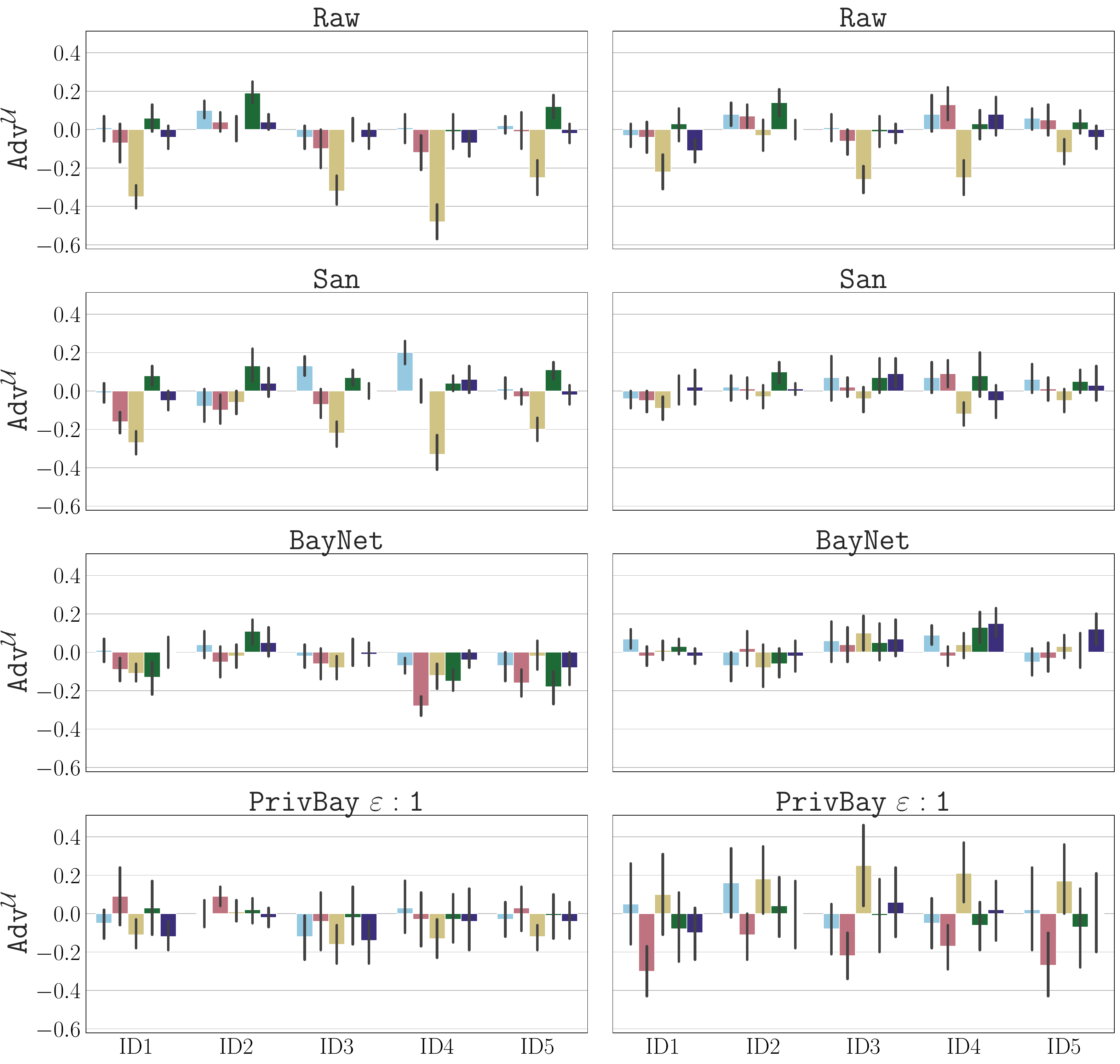}
	\caption{Utility advantage for $5$ test (indicated on the x-axis) and $5$ target (indicated by the colour of each bar) record pairs for predictions on attribute \texttt{RiskMortality} from the \Texas\ dataset for $\sizeR=1000$~(\emph{left}) and $\sizeR=5000$(~\emph{right})}
	\label{fig:util_adv}
\end{figure}

\parabf{Privacy through suppression.} We repeatedly ran the utility game and computed the utility advantage for each of the target-test record pairs. We show in \autoref{fig:util_adv} the utility advantage for a prediction task on attribute \texttt{RiskMortality} for two different dataset sizes ($n=1000$~\emph{left} and $n=5000$~\emph{right}). The colour of each bar indicates the target record $\target$ chosen by $\advGame$ and bars are grouped by test record $\test$ along the x-axis. For the differentially private models, we present data for a $\PrivBay$ model trained with $\varepsilon = 1.0$, the highest $\varepsilon$ value under which all targets still receive robust protection against linkage attacks.

Independent of a model's training set size, sanitised datasets tend to reproduce patterns found in the raw data. If a target's presence in the raw data has a significant (negative or positive) impact on a test record's prediction accuracy, this advantage is retained in the sanitised data. Synthetic data sampled from either model ($\BayNet$ and $\PrivBay~\varepsilon: 1.0$) exhibits entirely different results.
For instance, the presence of target record \tikzbar{safeYellow} in the raw training data leads to a negative prediction advantage with $|\advantageU| \geq 0.3$ for four out of the five test records with $n=1000$ (\autoref{fig:util_adv}~\emph{left}). The same holds true when the classifier is trained on data produced by $\San$. With training on synthetic data produced by $\BayNet$ and $\PrivBay$, the test records' advantage from the target's presence vanishes (maximum advantage of $|\advantageU| \leq 0.15$ and $\advantageU| \leq 0.06$).

This indicates that while row-level sanitisation preserves the statistical signals of outliers, and their potential positive impact on the prediction accuracy of similar test records, synthetic data produced by the two models evaluated here does not retain the targets' unique influence. In the case of $\PrivBay$, this is expected: The model's differential privacy guarantee ensures that the addition of a single record to its training set does not affect the model's output distribution by more than the defined $\varepsilon$-bound. This is also reflected in the \emph{increased privacy gain} of these targets under $\PrivBay~\varepsilon: 1.0$ compared to $\San$ shown in \autoref{fig:mia_san}. Our utility evaluation reveals that this \emph{gain in privacy does not come for free}.
As an example, the target marked \tikzcross{safeGreen} receives a low gain in privacy from sanitised data publishing with $\PG \leq 0.3$ under all three feature sets. Accordingly, the high positive impact the target's presence in the raw data has on test record ID2 ($\advantageU = 0.19$ shown in \autoref{fig:util_adv}) is preserved under sanitisation. Synthetic data sampled from $\PrivBay~\varepsilon: 1.0$ increases the target's privacy gain to $\PG \geq 0.77$ but simultaneously reduces the test record's advantage to $\advantageU = 0.02$. The same patterns can be found for other target records and dataset sizes.     

\parabf{Conclusions.} The inherent tradeoff between the privacy and utility of high-dimensional data releases has been shown many times~\cite{NarayananS08 , PyrgelisTD18, JayaramanE19}. We present empirical evidence that synthetic data publishing is subject to the same limitations, and might even provide less beneficial tradeoffs.
Synthetic data that protects outliers from linkage attacks does so at a cost in utility for test records from similar minority subgroups. Differentially private data releases provide more robust protection, 
but inadvertently suppress the statistical signal of the protected records. This decreases the utility of the released data and prevents statistical models from learning patterns about certain target groups, potentially increasing the bias and unfairness of data-driven decision making~\cite{BagdasaryanPS19, ChengSD21}.   

\section{Key takeaways}


 Synthetic data is often portrayed as a silver-bullet solution to privacy-preserving data sharing that provides a higher gain in privacy at a lower cost in utility than traditional anonymisation techniques~\cite{DrechslerR10, ChoiBMDSS17, XuSCV19 , YaleDDGPB19, YaleDDGPB19-A, BellovinDR19, ArnoldN20}. In this paper, we quantitatively assess this claim and demonstrate that it rarely holds true. The basic tradeoff between utility and privacy for high-dimensional data sharing remains: If a synthetic dataset preserves the characteristics of the original data with high accuracy, and hence retains data utility for the use cases it is advertised for, it simultaneously enables adversaries to extract sensitive information about individuals.
A high gain in privacy through any of the anonymisation mechanisms we evaluated can only be achieved if the published synthetic or sanitised version of the original data does not carry through the signal of individual records in the raw data and in effect suppresses their record.  

This is not an unexpected finding. Previous studies on the privacy-utility tradeoff of row-level data sharing~\cite{NarayananS08}, predictive ML models~\cite{JayaramanE19, CarliniDG20}, and aggregate query releases~\cite{PyrgelisTD18} all show that in order to protect privacy it is necessary to strictly limit information leakage about individuals. Our study demonstrates that synthetic data sharing does not magically resolve this tension but suffers from the same limitations.

Our findings not only show that synthetic data is subject to the same tradeoffs as previous anonymisation techniques but also demonstrate that the privacy gain of synthetic data publishing is \emph{highly unpredictable}. Because it is not possible to predict which data features a generative model will preserve, it is neither possible to anticipate the minimum gain in privacy from synthetic data publishing nor its utility loss. In comparison to deterministic sanitisation techniques, synthetic data does not allow data holders to provide transparency about what information will be omitted in the published dataset and what information will be retained.

We conclude that synthetic data does not provide a better tradeoff between privacy and utility than traditional row-level sanitisation, especially for data use cases that focus on the analysis of outlier signals, such as financial fraud or medical anomaly detection~\cite{TuckerWRM20, MostlyAI_Fraud}. Generative models with formal privacy guarantees reduce private information leakage with respect to anonymisation but do not preserve the fine-grained statistical patterns needed for outlier analysis.
Our evaluation further shows that, even for use cases that focus on aggregate insights, synthetic datasets with a high privacy gain can suffer from a significant utility loss and can lead to false conclusions. 
Even synthetic datasets that do preserve the desired statistics still present a noisier summary of the data than traditional privacy-preserving query release mechanisms~\cite{DworkMNS06} due to the additional uncertainty introduced by the output sampling process.

{\small
\para{\textbf{Acknowledgments.}} We would like to thank Jon Ullman, Aloni Cohen, Kobbi Nissim, and Salil Vadhan for their valuable feedback on earlier versions of this work, Laurent Girod for his support in open-sourcing our code, and the anonymous reviewers and our shepherd Takao Murakami whose input helped to further improve this paper. 
This work was partially funded by the Swiss National Science Foundation with grant 200021-188824.      
}

{\small \bibliography{mybib}}
\bibliographystyle{plain}

\vspace{-4mm}
\section{Appendix}\label{appendix}

\subsection{Notation}\label{ap:notation}
\autoref{tab:notation} summarises frequently used notation.

\begin{table}[h!]
	\centering
	\caption{Frequently used notation}
	\begin{tabular}{ll}
		\label{tab:notation}
		\textbf{Symbol} & \textbf{Meaning} \\
		\hline
		$\pop$ & Population, a collection of data records\\
		$r_i$ & Real-valued data attribute \\
		$\record = (r_1, \cdots, r_\nAttr)$ & Record, a real vector of $\nAttr$ attributes \\
		$R = (\record_1, \cdots, \record_\sizeR)$ & Raw dataset, a collection of $\sizeR$ records\\
		$\distPop = \Prob{\record}$ & High-dimensional joint probability\\
									& distribution over the data domain\\
		$\distR = \Prob{\record | R}$ & Joint probability distribution\\
									& induced by dataset $R$ \\
		$\GMtrained{R} \sim \trainGM{R}$ & Generative model obtained running \\
								& training algorithm $\trainGM{R}$ on dataset $R$ \\
		$\distGM{R}$ & Approximation of the joint distribution\\
						& of dataset $R$ through model $\GMtrained{R}$\\
		$S = (\synrecord_1, \cdots, \synrecord_\sizeS)$ & Synthetic dataset, collection of\\
														& $\sizeS$ records sampled from\\
														& a generative model $\GMtrained{R}$\\
		\\
		\hline
		\vspace*{-2mm}
		\\
		$\adv$ & Privacy adversary\\
		$\advMIA$ & Membership inference adversary\\
		$\advAI$ & Attribute inference adversary\\
		$\prior$ & Adversary's prior knowledge\\
		$\chGame$ & Challenger\\
		$\target = (r_1, \cdots, r_\nAttr)$ & Target record chosen by the adversary\\
		$\targetpartial = (r_1, \cdots, r_{\nAttr-1})$ & Partial target record\\
														& known to the adversary\\
														& excluding a sensitive attribute $r_s$\\
		$\secret$ & Membership secret of the target record\\
		$\secrethat$ & Guess about the target's secret $\secret$
	\end{tabular}
\end{table}

\vspace*{-2mm}
\subsection{Framework implementation}\label{ap:implementation}


We implemented the evaluation framework as a Python library~\cite{SynPrivEval}.
The library has two main classes: $\GenerativeModel$s and $\PrivacyAttack$s. For both classes we define a parent class that determines the core functionality that objects of the class need to implement.\\
$\GenerativeModel$ provides two main functions. $\mathtt{GM.fit()}$ is called with a raw dataset $\mathtt{R}$ as input and implements the model's training procedure. $\mathtt{GM.sample(\sizeS)}$ generates a synthetic dataset $\mathtt{S}$ of size $\mathtt{\sizeS}$ corresponding to $S \sim \distGM{R}^\sizeS$. The library enables easy integration of existing model training procedures. $\mathtt{GM.fit()}$ simply wraps any existing training algorithm and exposes the appropriate API endpoints.\\
$\PrivacyAttack$ objects have two functions: $\mathtt{PA.train}$ and $\mathtt{PA.attack}$. $\mathtt{PA.train(\target, BK)}$ trains the attack for a specific target record $\mathtt{\target}$ on background knowledge $\mathtt{BK}$. 
$\mathtt{PA.attack(S)}$, takes a dataset $\mathtt{S}$ and outputs a guess about a secret value. In our implementation, we instantiate $\PrivacyAttack$ with two attacks, a membership inference adversary and an attribute inference attack. 
The library also includes procedures to estimate the privacy gain of synthetic and sanitised data publishing.

\parabf{Generative model and feature set parametrisation.} We integrated five existing models into our Python library.
Each of the models has a set of model hyper-parameters that can be adjusted to fit the input data. In \autoref{tab:modelparams} we list the parameter values for each model and dataset used in our experiments.

\begin{table}
	\caption{Generative Model Hyperparameters}
	\label{tab:modelparams}
	{	\small

	\begin{tabular}{l|c|c|c|c|c}
		\toprule
			& $\IndHist$ & \multicolumn{2}{c}{$\BayNet$} & \multicolumn{2}{c}{$\PrivBay$}\\
			& $\texttt{nbins}$ & $\texttt{nbins}$ & $\texttt{degree}$ & $\texttt{nbins}$ & $\texttt{degree}$\\
		\midrule
		\Adult & $45$ & $45$ & $1$ & $45$ & $1$\\
		\Texas & $25$ & $25$ & $1$ & $25$ & $1$\\
	\end{tabular}
		\begin{tabular}{l|c|c|c|c}
		\midrule
			& \multicolumn{4}{c}{$\CTGAN$}\\
			& $\texttt{embeddings}$ & $\texttt{gen\_dim}$ & $\texttt{dis\_dim}$ & $\texttt{l2scale}$\\
		\midrule
		\Adult & $128$ & $(256, 256)$ & $(256, 256)$ & $10^{-6}$ \\
		\Texas & $128$ & $(256, 256)$ & $(256, 256)$ & $10^{-6}$ \\
		\bottomrule
	\end{tabular}
	}
	\vspace{-2mm}
\end{table}

%
%
%
We implement the adversary's feature sets as feature extraction objects $\mathtt{FeatureSet}$. Each $\mathtt{FeatureSet}$ takes in a synthetic dataset $S$ of size $\sizeS \times \nAttr$ and outputs a vector of size $\nFeatures \times 1$. Our library includes the following feature sets:

\para{$\Fnaive$.} The naive feature set computes the mean, median, and variance of each numerical attribute and encodes the number of distinct categories plus the most and least frequent category for each categorical attribute.

\para{$\Fhist$.} The histogram feature set computes the marginal distribution of each data attribute. Numerical attributes are binned with configurable bin size and frequency counts are computed for categorical attributes. The number of bins per attribute is configured for each dataset independently. In our experiments, we set the number of bins to $45$ and $25$ for the $\Adult$ and $\Texas$ dataset, respectively.

\para{$\Fcorr$.} The correlations feature set encodes pairwise attribute correlations. Categorical attributes are dummy-encoded before computing the pairwise correlation matrix.
%
The $\Fhist$ and $\Fcorr$ feature sets include a pre-processing step in which continuous columns are binned. The number of bins is a configurable parameter that can be adjusted to fit the input data. In our experiments, we set the number of bins to $45$ and $25$ for the $\Adult$ and $\Texas$ dataset, respectively.

\subsection{Datasets}\label{ap:datasets}
We include two tabular datasets, commonly used in the machine learning (ML) literature, in our experimental evaluation. Tabular datasets are the most relevant data type in the synthetic data publishing case. One datasets contains financial data the other one health data: 
 
\para{\Adult~\cite{KohaviB13}}. The Adult dataset contains information from 45,222 individuals extracted from the 1994 US Census database. Each entry consists of 15 attributes among which 6 are continuous attributes and 9 are categorical attributes.

\para{\Texas~\cite{Texas}}. The Texas Hospital Discharge dataset is a large public use data file provided by the Texas Department of State Health Services. The dataset we use consists of 50,000 records uniformly sampled from a pre-processed data file that contains patient records from the year $2013$. We retain 18 data attributes of which 11 are categorical and 7 continuous.
%
\vspace*{-3mm}
\subsection{Attribute inference}\label{ap:attribute}
We formalise the risk of attribute inference as a prediction problem in which an attacker learns to predict the value of an unknown sensitive attribute from a set of known attributes given access to a raw, sanitised, or synthetic dataset which we denote as $X$.
The adversary splits the dataset $X$ into two parts: A feature matrix $\tilde{X}$ that contains the values for all attributes known to the adversary and a vector $\bm{x}_s$ with the corresponding sensitive attribute values.

If the attribute targeted by the attack is a continuous, real-valued attribute $\sensitive \in \mathbb{R}$, we model attribute inference as a linear regression problem.
The linear regression attack models the relationship between the sensitive attribute values in $\bm{x}_s$ and the attributes in $\tilde{X}$ as a a linear relationship with coefficients $\coeffs{X}$, and treats the records in $X$ as i.i.d. samples:

\vspace*{-2mm}
\begin{equation}\label{eq:linreg}
	\bm{x}_s = \tilde{X} \coeffs{X} + \epsilon, \; \epsilon_i \sim \mathcal{N}\left( 0, \sigma^2 \right)\,,
\end{equation}

During training, the adversary takes the dataset $X$, splits it into a feature matrix $\tilde{X}$ and target variable $\bm{x}_s$, and uses maximum likelihood estimation to obtain a set of regression coefficients $\coeffs{X} = \max_{\coeffs{ }} \Prob{\bm{x}_s | \tilde{X}, \coeffs{}}$. 

The simplicity of the model enables us to analytically derive the attacker's posterior distribution over the target's secret given access to dataset $X$ 
	$\Prob{ \sensitivehat | X, \targetpartial} = \mathcal{N}\left[ \targetpartial \coeffs{X}, \hat{\sigma}^2_X \right]$, 
with variance $\hat{\sigma}^2_X = \frac{1}{\sizeR-(\nAttr-1)}\sum^\sizeR_{i=1}\left( x_s^i - \tilde{\bm{x}}^i \coeffs{X} \right)^2$.

\vspace{-2mm}
\subsection{Privacy-utility tradeoff}\label{ap:utility}
\begin{figure}
	\includegraphics[width=.5\textwidth]{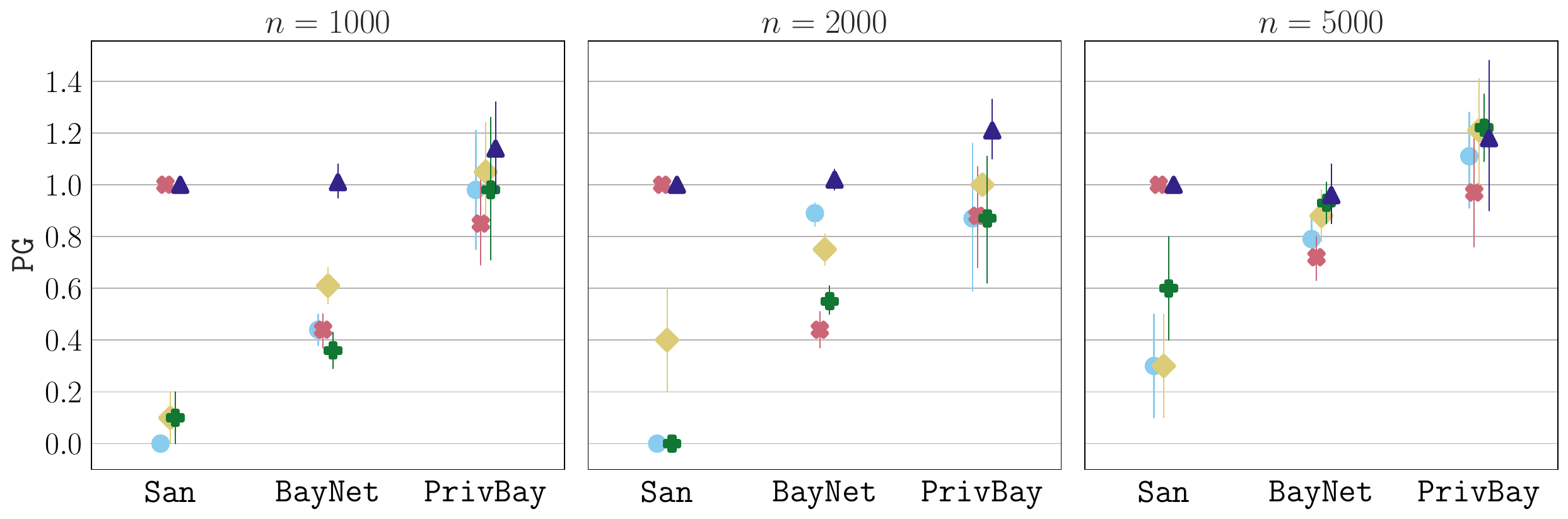}
	\caption{Per-record privacy gain for five outlier records for the \Texas\ dataset under an attack using the $\Fnaive$ feature set.}
	\label{fig:mia_sizes}
	\vspace*{-4mm}
\end{figure}

\autoref{fig:mia_sizes} shows the privacy gain for five outlier targets from the \Texas\ dataset for varying dataset sizes.
 
\end{document}